\documentclass[preprint,12pt]{elsarticle}

\usepackage{amssymb}
\usepackage{amsmath}
\usepackage[normalem]{ulem}

\usepackage{framed,multirow}
\usepackage{algorithmic}
\usepackage{algorithm}

\usepackage{comment}
\usepackage{url}
\usepackage{xcolor}
\usepackage{hyperref}
\usepackage{float}
\usepackage{amsmath}
\definecolor{newcolor}{rgb}{.8,.349,.1}
\usepackage{hyperref}
\usepackage{soul}
\usepackage{graphicx}
\usepackage{subfigure} % 加载 subfigure 宏包
\definecolor{darkblue}{rgb}{0.0, 0.0, 0.55}

\journal{Neurocomputing}

\begin{document}

\begin{frontmatter}



\title{Skeleton-OOD: An End-to-End Skeleton-Based Model for Robust Out-of-Distribution Human Action Detection} %% Article title


\author[2]{Jing Xu} %% Author name
%\ead{jing.xu1@monash.edu}
\author[3]{Anqi Zhu}
%\ead{azzh1@student.unimelb.edu.au}
\author[2]{Jingyu Lin}
%\ead{jingyu.lin@monash.edu} 
\author[2]{Qiuhong Ke}
%\ead{qiuhong.ke@monash.edu}
\author[2]{Cunjian Chen\corref{cor1}}
\cortext[cor1]{Corresponding author: 
  Tel.: +86 1-517-303-2723; }
\ead{cunjian.chen@monash.edu}
% \affiliation[1]{organization={Monash Suzhou Research Institute},
%              % addressline={},
%              city={Suzhou},
%              postcode={215000},
%              state={Jiangsu},
%              country={China}}
\affiliation[2]{organization={Department of Data Science and AI, Monash University},%Department and Organization
            city={Melbourne},
            postcode={3800}, 
            state={Victoria},
            country={Australia}}
\affiliation[3]{organization={Department of Computing and Information Systems, University of Melbourne},%Department and Organization
            city={Melbourne},
            postcode={3052}, 
            state={Victoria},
            country={Australia}}

%% Abstract
\begin{abstract}
Human action recognition is crucial in computer vision systems. However, in real-world scenarios, human actions often fall outside the distribution of training data, requiring a model to both recognize in-distribution (ID) actions and reject out-of-distribution (OOD) ones. Despite its importance, there has been limited research on OOD detection in human actions. Existing works on OOD detection mainly focus on image data with RGB structure, and many methods are post-hoc in nature. While these methods are convenient and computationally efficient, they often lack sufficient accuracy, fail to consider the exposure of OOD samples, and ignore the application in skeleton structure data. To address these challenges, we propose a novel end-to-end skeleton-based model called Skeleton-OOD, which is committed to improving the effectiveness of OOD tasks while ensuring the accuracy of ID recognition. Through extensive experiments conducted on NTU-RGB+D 60, NTU-RGB+D 120, and Kinetics-400 datasets, Skeleton-OOD demonstrates the superior performance of our proposed approach compared to state-of-the-art methods. Our findings underscore the effectiveness of classic OOD detection techniques in the context of skeleton-based action recognition tasks, offering promising avenues for future research in this field. Code is available at \url{https://github.com/YilliaJing/Skeleton-OOD.git}.
\end{abstract}

% \includegraphics[width=\textwidth]{figs/framework2.png}

% \begin{highlights}
% \item Propose an end-to-end skeleton-based OOD detection model that can address overconfidence.
% \item Introduce an attention-fused block to enhance OOD detection while maintaining ID classification.
% \item Design an energy-based loss function to improve score differentiation between ID and OOD samples.
% \item Outperform baselines in the human action OOD detection task by a noticeable margin.
% \end{highlights}

% %% Keywords
% \begin{keyword}
% %% keywords here, in the form: keyword \sep keyword
% Action Recognition \sep Out-of-Distribution Detection \sep Graph Neural Networks \sep Energy-based Theory
% %% PACS codes here, in the form: \PACS code \sep code

% %% MSC codes here, in the form: \MSC code \sep code
% %% or \MSC[2008] code \sep code (2000 is the default)

% \end{keyword}

\end{frontmatter}



\section{Introduction}
\label{introduction}
%% Labels are used to cross-reference an item using \ref command.

Human-centric visual understanding enhances the development of intelligent machines, enabling humans to accomplish greater tasks and improve their quality of life. Deep learning-based models have shown remarkable performance in supervised human action classification tasks. In skeleton-based human action datasets, this task can be solved as a typical spatio-temporal modeling problem \citep{yan2018spatial, chen2021channel, cheng2020decoupling}. Although these methods have achieved high accuracy in supervised classification problems, they still lack consideration for out-of-distribution (OOD) issues. An important assumption in these approaches is that the distribution of the training set is similar to that of the test set. However, this can not always be achieved in practice. When straightly applying these action recognition models to recognize samples outside the distribution of the training dataset, they may exhibit overconfidence by selecting a category from the known distribution \citep{yu2024exploring}. Hence, it is essential for such models to be aware of the boundaries of learning for safety concerns \citep{nalisnick2018deep}, ensuring the robustness of computer vision systems. How to deal with this OOD detection problem has attracted widespread attention in fields such as autonomous driving \citep{geiger2012we} and medical image analysis \citep{schlegl2017unsupervised}. 

%泛领域OOD：目前发展到什么地步，是否贴合Skeleton数据的特征处理需求→自己的方法解决了这个问题

Works in OOD detection focus on OOD samples caused by semantic shift, aiming to determine whether a sample belongs to a category included in the training dataset. This task presents two challenges: quantifying the differences between in-distribution (ID) and OOD samples and distinguishing them based on their expressions. For the first challenge, one typical approach is score measuring methods \citep{hendrycks17baseline}, such as the softmax confidence score \citep{hendrycks17baseline}. Inputs with low softmax confidence scores are classified as OOD. However, neural networks can produce arbitrarily high softmax confidence scores for inputs far from the training data \citep{nguyen2015deep}. To address this, the energy score is proposed for OOD detection. It can be derived from a purely discriminative classification model without explicitly relying on a density estimator. Experiments have shown that it outperforms softmax-based score measuring methods.
% Energy函数：相比其他OOD有什么区别，如何发展→如何用来解决Skeleton OOD的问题

Based on such score measuring methods, a crucial issue is how to maximize the difference in score distributions between OOD and ID samples. To achieve this, difference amplification methods \citep{sun2021react, zhu2022boosting} have been designed. However, these methods are mostly implemented in a post-hoc manner, meaning the model itself cannot inherently distinguish OOD samples. This does not fundamentally solve the problem of `letting the model know what it knows'. Furthermore, these methods were originally designed for OOD detection in RGB-structured datasets. Their effectiveness in handling action recognition problems with skeleton structure data has not yet been investigated.

Furthermore, methods like \cite{wu2023gnnsafe, koo2024generalized} have attempted to achieve this goal by introducing OOD-related information into the training stage. However, in practice, we cannot measure the specific categories of OOD. Similarly, when the OOD category itself expands, the overconfidence problem still remains. In summary, while these methods significantly improve the accuracy of OOD recognition, achieving this goal using an end-to-end framework without prior OOD information remains challenging.

To address the limits and challenges mentioned above, we propose an end-to-end model called Skeleton-OOD to deal with the out-of-distribution detection problem for skeleton-based action recognition. Considering that the state-of-the-art feature extractors for supervised ID classification are well-developed, we focus on exploiting the extracted features to extend their adaptability for OOD detection while maintaining robust prediction performance for ID classification. We input graphs, joints, and temporal sequences into the spatial-temporal feature extractor to obtain feature embeddings as output. Based on this, we first creatively applied a feature activation module that utilizes ASH \citep{djurisic2022ash} to project the original features for fitting OOD detection in the training stage. Specifically, it filters useless information and amplifies the key feature dimensions to generate a more sensitive and accurate representation of abnormal detections. However, in its original work, the activation method was only used in a post hoc manner after the trained model, and the energy score was only used for judgment. Then, regarding this as an orthogonal aggregation to the original features, we innovatively designed a feature fusion module, which integrates the original backbone outputs and the OOD-targeted activated features to enable the model to preserve a strong recognition ability for ID data while also nicely discriminating the potential OOD existences. To support the framework with end-to-end training, we suggested a meticulously designed learning loss that combines the energy function and cross-entropy information. This approach aims to improve the model's understanding of the energy score distribution in ID data, thereby further benefiting the accurate detection of OOD samples.

Our main contributions are summarized as follows:

\begin{itemize}
    \item We propose Skeleton-OOD, an end-to-end skeleton-based OOD detection model that addresses overconfidence by solely training on ID data.
    
    \item The attention-based feature fusion block is designed to enhance OOD detection accuracy while preserving classification ability for ID classes.
    
    \item We design an energy-based loss function, which can improve score differentiation between ID and OOD samples, thus effectively maximizing their distribution separations.
    
    \item The proposed model Skeleton-OOD outperforms baselines on NTU-RGB+D 60, NTU-RGB+D 120, and Kinetics-400 datasets in the human action OOD detection task.
\end{itemize}

The rest of the paper is organized as follows: Section \ref{relate} explains basic knowledge of skeleton-based action recognition tasks with GCNs and the theory of why energy score is most widely used in OOD detection. Section \ref{methodd} introduces our proposed method in detail, including the proposed framework, feature fusion block, and loss function design. Section \ref{experiments} mainly discusses the results of experiments and thus shows the effectiveness and usability of our proposed model. We offer conclusions, limitations, and future works in Section \ref{conclusion}.

\section{Related works}
\label{relate}

\subsection{GCNs for Skeleton-based Action Recognition}
\label{subsection21}

Human action recognition tasks are a typical problem in computer vision, which have wide applications in the real world, including human-computer interaction \citep{nikamambekar2016} and video surveillance \citep{jiang2015human}. Skeleton-based methods have their unique advantages. The skeleton structure can provide more accurate node-level information and ignore the influence of background noise, thus improving the prediction accuracy of recognition.

In the early phases of skeleton-based action recognition using deep learning methods, convolutional neural networks (CNNs) \citep{cheron2015p,liu2017enhanced,simonyan2014two}, and recurrent neural networks (RNNs) \citep{wang2017modeling,liu2017global} were commonly utilized. However, these approaches had limitations as they didn't fully leverage the structural arrangement of the joints.

Subsequently, researchers built spatio-temporal architectures to handle this problem. Many works use handcrafted physically connected (PC) edges among human skeletons to extract spatial features \citep{chen2021channel, cheng2020decoupling}, and GCNs can handle this type of graph-structured data very well. The graph defined on the human skeleton is denoted as $G(V, E)$, with $V$ representing the joint group and $E$ representing the edge group. The representation of 3D time-series skeletal data is denoted as $X\in\mathbb{R}^{c\times t\times |V|}$, where $|V|$ represents the number of joint nodes, $c$ is the number of channels, and $t$ denotes the temporal window size. The operation of GCN with an input feature map $X$ can be described as follows:

\begin{equation}
\mathbf{F}_{\text {out }}= \mathbf{A} \mathbf{X} \boldsymbol{\Theta} .
\end{equation}

Here, $\boldsymbol{\Theta}$ denotes the pointwise convolution operation. The adjacency matrix $A$, which is initialized as $\mathbf{\Lambda}^{-\frac{1}{2}} \mathbf{A} \boldsymbol{\Lambda}^{-\frac{1}{2}} \in \mathbb{R}^{N_g \times |V| \times |V|}$, where $\Lambda$ is a diagonal matrix for normalization, and $N_g=3$ in experiments.

However, relying solely on PC edges for spatial relationships is insufficient, as it limits the receptive fields due to their heuristic and fixed nature \citep{tian2024multi}. Additionally, the varying contributions of each edge must be considered, as the importance of joints in different body parts varies for specific actions. To address these issues, Hierarchically Decomposed Graph Convolutional Networks (HD-GCN) \citep{lee2023hierarchically} were proposed. The graph convolution operation of HD-GCN is defined as:
% Based on that, several efforts have been made to model complex relationships between body joints. \cite{shi2019two} and \cite{ye2020dynamic} used adaptive graph and dynamic graph construction methods. They overcame the limitations by the natural links between body joints, and provided a method to represent the intrinsic topology. Despite the spatial relationships, temporal correlations between different action frames were also considered. \cite{yan2018spatial} first proposed a spatio-temporal framework to model human actions. Expanding the receptive field of the model in time and space is also a crucial challenge to address. 

\begin{equation}
\mathbf{F}_{\text {out }}=\sum_{g \in G} \mathbf{A}_g \mathbf{X} \boldsymbol{\Theta}_g ,
\label{gcn}
\end{equation}

\noindent where $G = \left\{{g_{pc}, g_{sl}, g_{fc}}\right\}$ denotes three graph subsets, and $g_{pc}$, $g_{sl}$, and $g_{fc}$ indicate physical connections, self-loops, and fully connected joint subsets, respectively. It represents three typical useful graph relationships to help explore spatial relationships between joints. The construction of these three types of subsets will be introduced in detail in section~\ref{subsec31}. With this operation, HD-GCN achieved an average Top-1 accuracy of 93.9\% across three experimental datasets. 

\subsection{Methods for out-of-distribution detection }
\label{subsection22}
Research on out-of-distribution detection can be mainly categorized into two ways. First, we can define a scoring function that maps each input point to a single scalar, such that in-distribution and out-of-distribution data will have different distributions. Thus, we can identify the out-of-distribution samples by calculating this score. There are many works about how to define this kind of score for pre-trained neural networks. \cite{hendrycks17baseline} proposed the maximum predicted softmax probability (MSP). After that, \cite{wu2023gnnsafe} proved that the log of the MSP was equivalent to a special case of the free energy score. This indicates that there are some cases in ID and OOD data with similar MSP values but different energy scores. As a result, energy score has become one of the most widely used score measuring methods.

Secondly, we can distinguish ID and OOD by trying to modify the feature captured by a pre-trained neural network in a post-hoc way, which is usually applied to feature activations. Numerous experiments have shown that different feature activation methods can help differentiate ID and OOD. \cite{sun2021react} proposed rectified activation (ReAct) after observing that OOD data can trigger unit activation patterns that were significantly different from ID data in the penultimate layer of the model. Performing truncation can help drastically improve the separation of ID and OOD data. \cite{zhu2022boosting} found that some operations similar to Batch Normalization might increase the difference in scores (e.g., MSP and energy score) between ID and OOD data. DICE \citep{sun2022dice} was proposed to address the shortcomings of redundant information expressed in the high space of neural networks. It attempted to define weights (i.e., weight × activation) to sort and filter the nodes in the penultimate layer of the neural network for denoising. This further reduces the variance value of the ID and OOD data score distributions, making it easier to separate the peaks of the two distributions. Similar to the previous works, \cite{djurisic2022ash} removed the top-K elements (usually a large portion over 50\%) and adjusted the remaining (e.g. 10\%) activation values by scaling them up or straightly assigned them a constant number. It achieves state-of-art performance on several benchmarks of image classification. 

However, most of the first two kinds of methods are considered post-hoc. While they can reduce calculation costs than training another new specific OOD detection model, research by \cite{liu2023gen} showed that they were still less competitive than some works that utilized OOD information in the training stage, such as works by \cite{hendrycks2019scaling} and \cite{song2022rankfeat}. Additionally, these methods were initially designed for OOD detection in RGB-structured datasets. Their effectiveness in addressing action recognition problems with skeleton structure data has not yet been explored. In this work, we prove the validity of these methods on skeleton-based datasets and then take steps to solve these problems.

\subsection{Energy-based out-of-distribution detection}
\label{subsec23}
In deep neural networks, out-of-distribution detection distinguishes samples that deviate from the training distribution. In this work, we only focus on the standard OOD detection that concerns semantic shifts, in which OOD data is defined as test samples from semantic categories that were not seen during training. Ideally, the neural network should be able to reject such samples as OOD while maintaining strong classification performance on ID test samples belonging to seen training categories \citep{zhang2023openood}.  To achieve this goal, a common approach is to utilize the data density function, denoted as $p^{in}(x)$, and identify instances with low likelihood as OOD. Nevertheless, prior studies have demonstrated that density functions estimated by deep generative models are not consistently reliable for OOD detection \citep{nalisnick2018deep}.

Energy-based out-of-distribution detection employs the energy score for detection, where the difference of energies between ID and OOD facilitates separation. The energy score addresses a crucial issue associated with softmax confidence, which can lead to excessively high values for OOD examples \citep{hein2019relu}.

Let us define a discriminate neural network $f(x):\mathbb{R}^D\to\mathbb{R}^K$, which maps the input $x$ to $K$ real-valued numbers known as logits. Its energy score function $E(x,f)$ over $\mathbb{R}^D$ with softmax activation can be defined as follows:

\begin{equation}
    E(\mathbf{x} ; f)=-\varepsilon \cdot \log \sum_i^K e^{f_i(\mathbf{x}) / \varepsilon},
\label{energy}
\end{equation}
 
\noindent where $\varepsilon$ is the temperature parameter. It can also be proved that a model trained with negative log-likelihood (NLL) loss will push down energy for in-distribution data points and pull up for other labels \citep{lecun2006tutorial}. Since cross-entropy loss can be implemented as NLL loss after the logSoftmax function, it will also push down the energy score of ID data. In other words, the ID sample will have a larger negative energy score $-E(x;f)$ than the OOD sample.

\begin{figure*}[t]
  \centering
  \includegraphics[width=\textwidth]{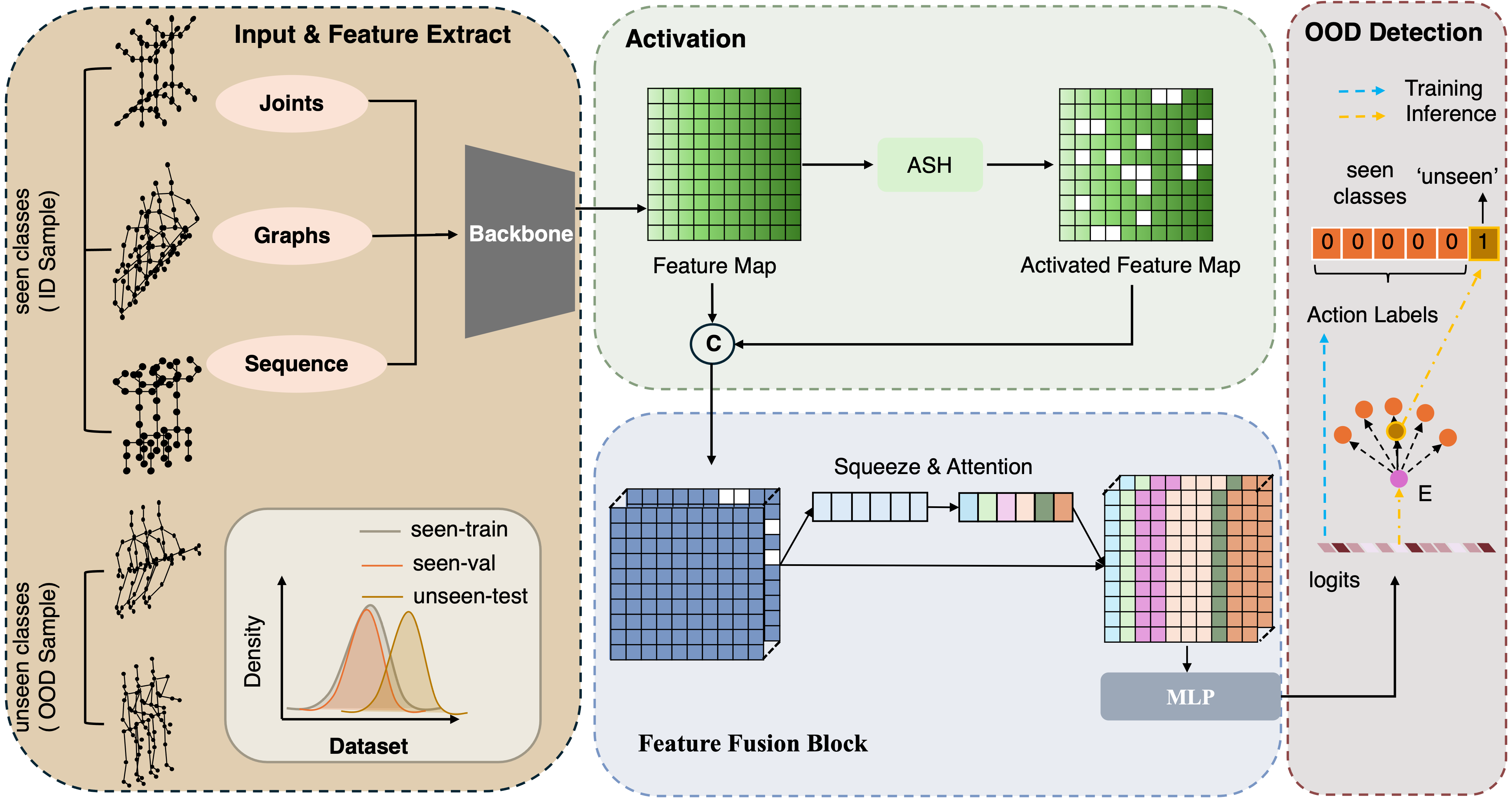}
  \caption{Framework of the proposed Skeleton-OOD model. The symbol \normalsize{\textcircled{\scriptsize{C}}}\normalsize \, represents the vector concatenate operation.}
  \label{pipeline}
\end{figure*}

\section{Method}
\label{methodd}
Our goal is to achieve accurate human action OOD detection through an end-to-end framework without any other subsequent fine-tuning or post-hoc operations. The whole framework of the proposed skeleton-based OOD human action detection model is shown in Figure~\ref{pipeline}.

It can be divided into three main steps of the end-to-end Skeleton-OOD framework: Firstly, a GCN-based backbone is employed to extract features from skeleton data. Secondly, The extracted features are then activated and fused with the original ones. The function of the feature fusion block is to improve the recognition ability of OOD samples while preserving the original classification performance of ID data. Finally, it is fed into the classifier, and OOD detection is performed based on the classifier output logits as shown in Figure~\ref{ood}. Besides, the whole model is trained by our newly designed energy-based loss function. In the following sections, we will provide a detailed discussion of each component.

\subsection{GCN-based feature extractor}\label{subsec31}

\begin{figure}[H]
  \centering
  \includegraphics[width=\textwidth, scale=0.6]{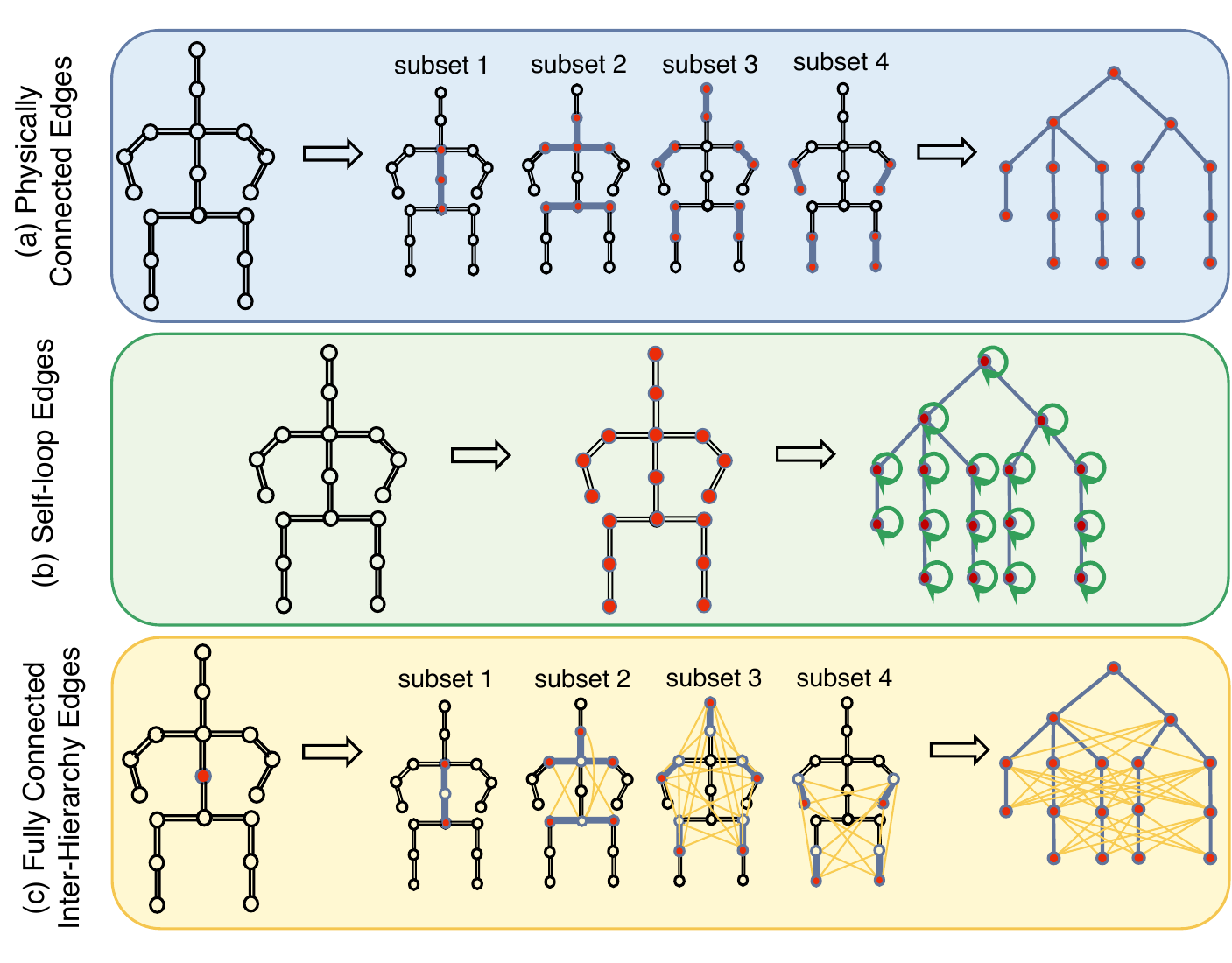}
  \caption{Description of the graph construction method. (a) Physically connected edges. The nodes and edges marked in each subset correspond to the nodes and edges of the two adjacent layers of nodes in the rightmost tree structure. (b) Self-loop edges (green arrow pointing to the joints themselves). We add self-loops of each joint to help the model take into account the characteristics of the joints themselves when conducting graph convolution. (c) Fully connected inter-hierarchy edges. We assume that the inter-hierarchy relationships in each subset can help to distinguish different actions. Thus, we add fully-connected edges in each subset as shown in yellow lines.}
  \label{graph}
\end{figure}

Since many GCN-based models have achieved great success in action recognition tasks, in this work we use existing models that perform best in ID data classification to extract embeddings from human skeleton data. It is an important step for a GCN-based model to construct the graph topology in advance. A topology can bring great convenience to the spatial learning of the model while incorporating prior information thus improving the physical consistency of the model. Following the previous work by \citep{lee2023hierarchically}, we define three types of edges for joint connections, while keeping the number of nodes unchanged (see Figure~\ref{graph}).

% 改
Let us define $G = \left\{{g_{pc}, g_{sl}, g_{fc}}\right\}$ as the three types of graph. The first type of graph $g_{pc}$ is physically connected edges, which are represented by blue lines between joints in Figure~\ref{graph} (a). It represents the natural physical relationships in skeleton data. The second type $g_{sl}$ is the self-loop, where nodes are connected to themselves. By incorporating self-loops, nodes are enabled to consider their own features during message passing in addition to relying on neighboring nodes' information, which helps stabilize the training process. Additionally, self-loops can be regarded as a regularization mechanism to prevent over-fitting, making them a commonly used method in graph neural network construction\citep{kipf2016semi}.

The third type of graph is a hierarchical fully connected graph $g_{fc}$. We employ a balanced tree approach to reconstruct the graph structure (as illustrated on the right side of Figure~\ref{graph}). Fully connected edges are constructed between adjacent nodes across two levels of the balanced tree, based on the hierarchical level of the nodes. This approach considers that the characteristics of actions may correlate with features across various levels of the body's hierarchical structure. Moreover, it expands the model's receptive field spatially in the feature space, enhancing the model's generalization ability. This approach has been widely proven to be effective in previous experiments. After that, for the input skeleton joints data $X$, graph subsets $G$ and temporal sequence $T$, the output feature maps $F$ are obtained as follows:

\begin{equation}
    F=f^{H D-G C N}(X, G, T).
\end{equation}

\subsection{Feature activation shaping and fusion}
\label{subsec32}

Following the previous work \citep{djurisic2022ash}, we utilize activation shaping (ASH) operation to filter out useless information in high-dimensional HD-GCN captured feature maps $F$, which can bring benefits to other downstream tasks (e.g. OOD detection). However, the difference is that we utilize it in the training stage to help the model itself distinguish OOD samples.

\begin{figure}[H]
  \centering
  \setlength{\abovecaptionskip}{0.cm}
  \includegraphics[scale=0.8]{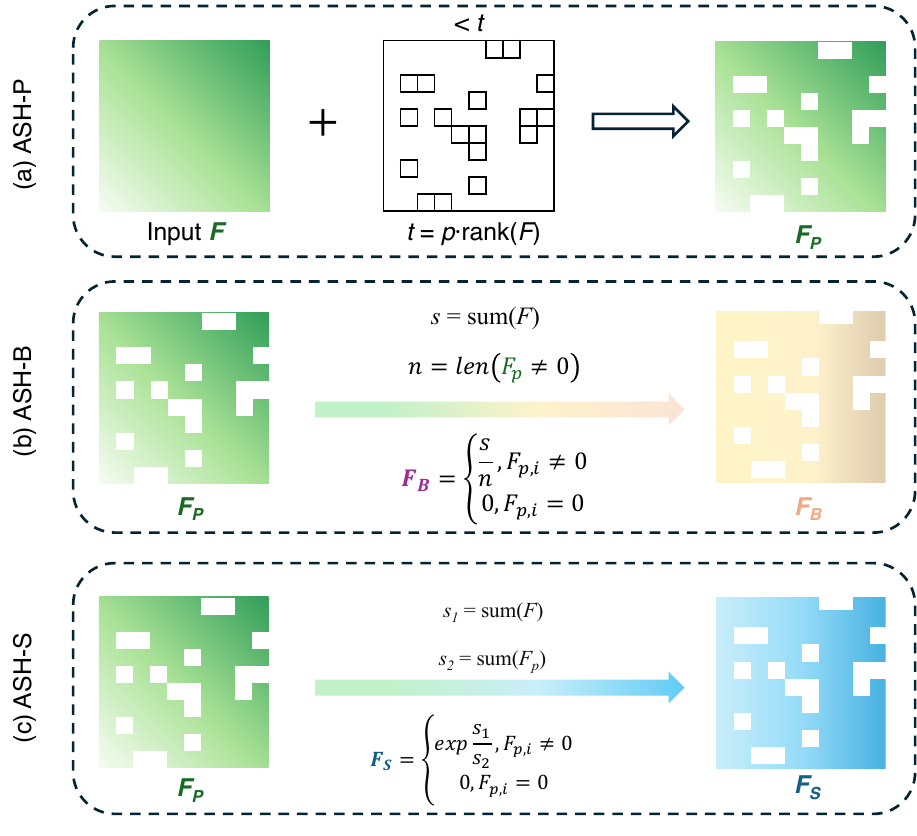}
  \caption{Process of Activation Shaping strategies. They prune the elements by pruning percentages and then conduct different strategies to assign values to the remaining elements.}
  \label{ash}
\end{figure}

There are three types of ASH strategies: ASH-P, ASH-B, and ASH-S. As shown in Figure~\ref{ash}, for the input feature maps $F$, ASH-P only sets values less than $t$ into zero, ASH-B resets the non-zeros by binarization, and ASH-S sets them with scaling. Details of these three strategies can be found in \ref{app1}. The difference among them lies in the normalization strategies after pruning. In the experimental part, we show the results of all these three strategies in training. After that, we retain both the output feature maps and the pruning features:

\begin{equation}
\hat{F}=\left[F, \text{ASH}(F)\right],
\label{concat}
\end{equation}

\noindent where ASH($F$) represents the results of any activation strategy acting on the feature maps $F$, and $[\cdot, \cdot]$ denotes the process to concatenate two vectors into a single one. The concatenated features undergo two main operations in the feature fusion process: one SE layer \citep{hu2018squeeze} primarily aimed at enhancing the model's sensitivity to channel features:

\begin{equation}
F^{SE}=\hat{F} \cdot \operatorname{ReLU}\left(W_{f c} \cdot \underbrace{\frac{\sum_i^{k_h} \sum_j^{k_w} \hat{F}}{k_h \times k_w}}_{\text{average pool}}+b\right),
\label{selayer}
\end{equation}

\noindent and the other MLP layer \citep{yu2022metaformer}:

\begin{equation}
F^{f u s e}=\operatorname{MLP}\left(F^{SE}\right).
\label{fuse}
\end{equation}

Here, $W_{fc}$ represents the learnable weights. At this stage, we have acquired the fused feature $F^{f u s e}$. Experiments show that it can improve the recognition ability of unseen classes while ensuring the classification accuracy of seen classes. 

\subsection{OOD detection using logits}\label{subsec33}
After the feature pruning and fusion process, the extracted features will be fed into a classifier:

\begin{equation}
logits=W_{f c}^{\prime} \cdot \text{drop\_out}\left(F^{f u s e}\right)+b^{\prime}.
\label{classify}
\end{equation}

\begin{figure}[H]
  \centering
  \setlength{\abovecaptionskip}{0.cm}
  \includegraphics[width=\linewidth,scale=0.7]{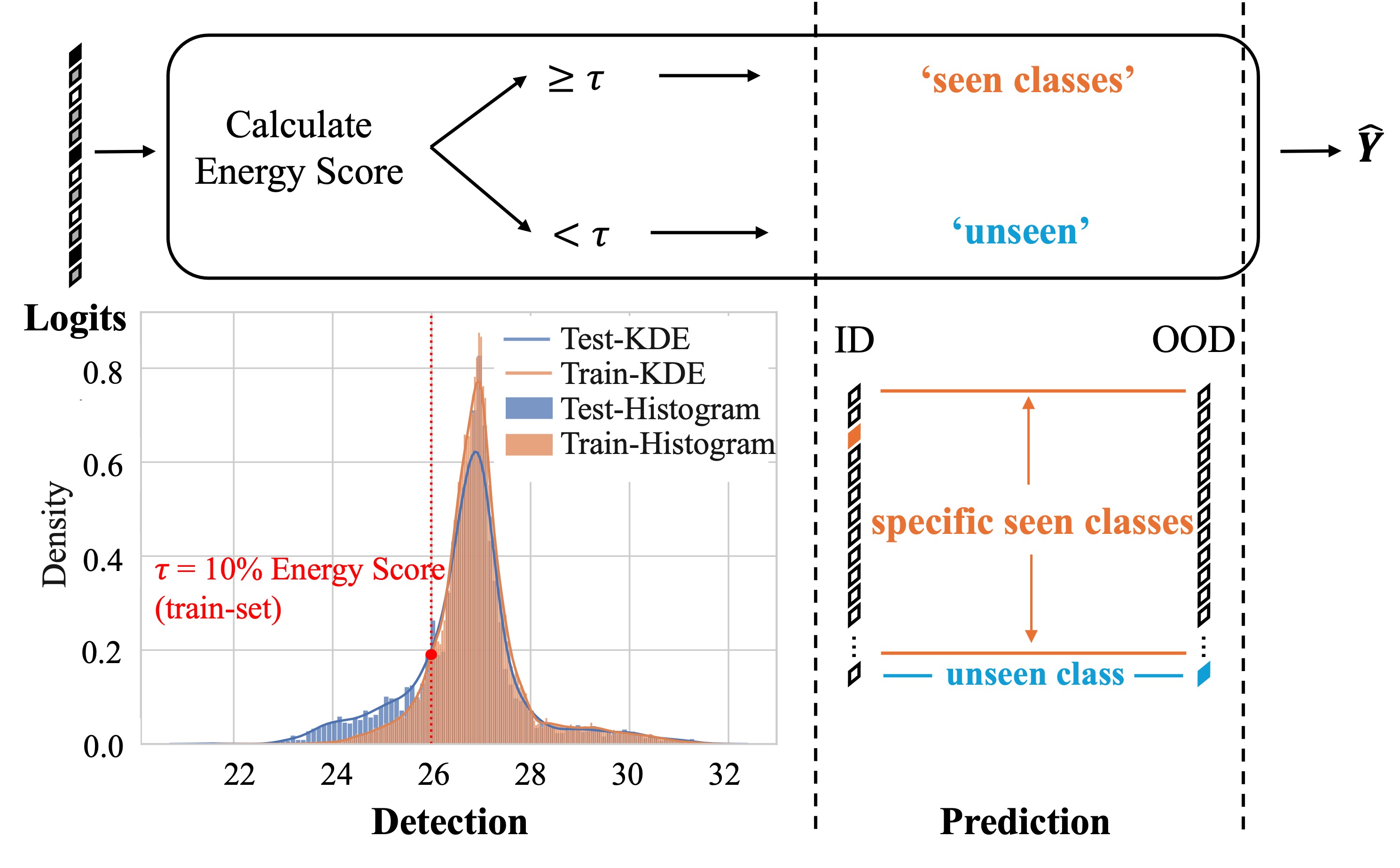}
  \caption{Explanation of OOD detection process. In the detection stage, the input is the logits from the previous feature extractor. Based on the Energy-based theory, the lower the score, the more likely the sample is the in-distribution.}
  \label{ood}
\end{figure}

Here Eqn. \ref{classify} shows the classification operation after data fusion, and $W_{f c}^{\prime}$ is the learnable weight. Then we get the logits ready for the downstream OOD detection task. The logits' dimension equals the number of seen classes plus one unseen class. The detailed process of OOD detection is illustrated in Figure~\ref{ood}.

For the input logits after the classifier, we measure the energy score of each sample using Eqn.~\ref{energy}: If the score is smaller than $\tau$, we will recognize it as ‘unseen’ for OOD and set the element value representing the probability that the sample belongs to the unknown category to 1. Otherwise, the probability of unseen will remain 0, and others (representing specific seen classes) will be the value between 0 to 1. The output of this stage will be multi-classes. The value of parameter $\tau$ is given based on the top 10\% of energy scores on the training set. The assumption behind this is that the energy score of OOD after training is less than the energy score of 90\% ID data. After that, we can get a vector $\hat{Y}$ that represents the probability that the sample belongs to each class.

\begin{algorithm}[H]
  \renewcommand{\algorithmicrequire}{\textbf{Input:}}
  \renewcommand{\algorithmicensure}{\textbf{Output:}}
      \caption{Skeleton-OOD Detection Algorithm}
      \label{alg}
      \begin{algorithmic}[1]
      \REQUIRE{
          Action skeleton data $X$; \\
          Graph structure $G = \left\{g_{id}, g_{cf}, g_{cp}\right\}$; \\
          Detection threshold $\tau$ from training stage; \\
          Algorithm running STAGE;
      }
      \ENSURE{
          Predict action label $\hat{Y}$ for inference or value of loss function $L$ in the training stage.
      }
      \STATE{
          $F=f^{HD-GCN}(X, G, T)$
      } \COMMENT{extract feature maps}
      \STATE{
          $\hat{F}=[F, ASH(F,p)]$
      }\COMMENT{feature fusion block}
      \STATE{
          $F^{SE}=SELayer(\hat{F})$  
      }\COMMENT{detailed in Eqn.~\ref{selayer}}
      \STATE{
          $F^{fuse}=MLP(F^{SE})$
      }
      \STATE{
          $logits=Classification(F^{fuse})$
      }\COMMENT{detailed in Eqn.~\ref{classify}}

      \IF{STAGE == `train'}
        \RETURN $L(logits)$   \COMMENT{detailed in Eqn.~\ref{energyloss}}
      \ELSE
        \STATE{
          $score=-E(logits)$
          }\COMMENT{calculate energy score}
          \IF{$score\geq \tau$} 
            \STATE $logits$[-1] = 1  \COMMENT{detect as OOD}
          \ENDIF
          \STATE $\hat{Y}=argmax(softmax(logits))$  
          \COMMENT{get label through the normalized one-hot vector}
          \RETURN $\hat{Y}$
          \STATE Calculate evaluation metric.
      \ENDIF
      \end{algorithmic}
\end{algorithm}

\subsection{Energy-based loss function}\label{subsec34}

As stated in section \ref{subsection22}, although the model trained with cross-entropy loss will lower the energy score value of OOD data, the gap between the two distributions may not always be optimal for detection. Therefore, we propose an energy-bounded learning objective and combine it with traditional cross-entropy loss:

\begin{equation}
\begin{split}
L=&-\sum_{logits_{in}}P(logits_{in}) \log f(logits_{in}) \\
&+ \alpha \cdot \mathbb{E}_{\left(logits_{in}, y\right) \sim \mathcal{D}_{\text {in }}^{\text {train }}}\left(\max \left(0, E\left(logits_{in}\right)-m_{\text {in }}\right)\right)^2.
\end{split}
\label{energyloss}
\end{equation}

The first part is the cross-entropy part, where $P(logits_{in})$ is the expected probability output, and $f(\cdot)$ represents the model output on in-distribution training data. The second part is energy-based, where $\mathcal{D}_{\text {in}}^{\text {train}}$ is the in-distribution training data, $E\left(logits_{in}\right)$ represents the energy score of in-distribution data, $m_{in}$ is the margin parameter which is set before can help the ID energy score distribution reduce the variance and close to this point, $\alpha$ is the tuning parameter that adjusts the proportion of energy part. With the help of this learning objective, the variance of ID data will be reduced. We can also ensure the model understands the energy score distribution of ID data,  thus indirectly having the ability to distinguish ID and OOD samples. The overall process can be found in Algorithm~\ref{alg}.

\section{Experiments}
\label{experiments}

To demonstrate the human action detection accuracy of our proposed model Skeleton-OOD, a series of experiments are conducted. In this section, we elucidate the intricate details of the experimental design and present a thorough analysis of the experimental results. First, we introduce three datasets and describe their data pre-processing methods. Subsequently, we compare our work with other existing OOD detection methods and identify the proposed model's strengths and weaknesses. Finally, we design three ablation studies to examine the effectiveness of individual modules.

\subsection{Dataset description}

\begin{itemize}
    \item \textbf{NTU-RGB+D 60 \citep{shahroudy2016ntu} \& 120 \citep{liu2019ntu}} \footnote{\href{https://rose1.ntu.edu.sg/dataset/actionRecognition/}{https://rose1.ntu.edu.sg/dataset/actionRecognition/}}: The NTU-RGB+D 60 dataset (NTU60), comprising 60 motion categories and 56,880 video samples, is extensively employed in diverse action recognition tasks. The NTU-RGB+D 120 dataset (NTU120) is an extension of the NTU-RGB+D 60 dataset, expanding the total number of human action categories to 120. Both datasets encompass five data modalities: 3D skeletons (body joints), masked depth maps, full-depth maps, RGB videos, and IR data. In this study, we exclusively utilize the 3D skeleton sequence data. The 3D skeleton data includes the 3D coordinates (X, Y, Z) of 25 body joints per frame for a maximum of two subjects. The actions in these datasets are categorized into three main groups: daily actions, interactive actions, and medical conditions. 5 classes are randomly selected as unseen from the NTU60 dataset and ten classes from the NTU120 dataset. Note that these unseen categories encompass samples from each of the three groups. Before splitting, we cleaned and preprocessed the raw dataset following the prior work \citep{lee2023hierarchically}.
    \item \textbf{Kinetics human action dataset} (Kinetics 400) \citep{kay2017kinetics}: It is, by far, the largest unconstrained action recognition dataset created by DeepMind. It covers 400 action classes, ranging from human-object interactions to human-human interactions, and also includes other complex interactions from YouTube videos. Since the Kinetics dataset provides only raw video clips without skeleton data, we adopt the skeleton-based action data processed by previous work STGCN \footnote{\href{https://github.com/yysijie/st-gcn/blob/master/OLD_README.md}{https://github.com/yysijie/st-gcn}} \citep{yan2018spatial}. The processed data sequences contain 18 joints and 2 subjects. We randomly select 33 action classes marked as ‘unseen’ for OOD samples, ensuring the same ratio as used in the NTU60 and NTU120 datasets. 
    \item \textbf{Dataset splitting protocol}: Both the training set and validation set consist of data with known labels. In this work, we create two types of test dataset: ID-only (randomly selected from known classes) and ID-OOD mix dataset, for the aim of testing the model's ability to classify ID samples and distinguishing OOD samples (all samples of unknown classes) from mixed samples. The seen class data in the validation set and the ID-only test set are consistent. The data volume ratio of the training set and ID-only test set/validation set in the known class data is 9:1. The number of samples in the train, validation, and test dataset is shown in Table \ref{datasets}. 
\end{itemize}

\begin{table}[H]
    \centering
    \caption{Number of samples in different data sets.} \small
    \begin{tabular}{c|c|c|c}
    \hline
    \textbf{Dataset}     & \textbf{Train}  & \textbf{Val}   & \textbf{Test (seen/unseen)} \\ \hline
    NTU60       & 46,898  & 5,184  & 5,184/4,730          \\ \hline
    NTU120      & 94,553  & 10,447 & 10,447/9,467         \\ \hline
    Kinetics400 & 216,441 & 24,049 & 24,049/19,742        \\ \hline
    \end{tabular}
    \label{datasets}
\end{table}

\subsection{Experiment details}

\subsubsection{Experiment settings}\label{subsubsec421}
\begin{itemize}
    \item \textbf{Implementation details}: We adopt HD-GCN \citep{lee2023hierarchically} as the backbone to extract graph-structured skeleton data as shown in Figure~\ref{pipeline}. The batch size of data is 64. The SGD optimizer is employed with a Nesterov momentum of 0.9 and a weight decay of 0.0004. Due to the differences in dataset size and data characteristics, we also used some different hyperparameters when training on different datasets: The hidden layer dimension of MLP is 400 for NTU and 576 for Kinetics. The learning rate is 0.001 for NTU60 and NTU120, and 0.01 for Kinetics400. The number of learning epochs is set to 100 for NTU60 and NTU120, and 150 for Kinetics400, with a warm-up strategy \citep{ he2016deep} applied to the first five epochs for more stable learning. The model used in this study contains a total of 1,815,060 parameters, leading to a trained model file size of approximately 9.3 MB. The training process for this model is efficient, with each epoch requiring about 7 minutes on the NTU60 dataset. The test on the NTU60 ID-OOD mix dataset (containing 9,914 samples) takes 14 seconds in total. All our experiments are conducted on NVIDIA GeForce RTX 3090, CUDA 11.6 + PyTorch 1.12.1.
    \item \textbf{Evaluation metric}: As depicted in Figure~\ref{ood}, we assess both the accuracy of detecting unseen samples and classifying seen samples. All samples under evaluation first undergo a threshold-based determination to discern whether their labels are unseen. If classified as `unseen', they are marked accordingly; otherwise, they are predicted to belong to known categories. Regarding the unseen detection task, it can be viewed as a binary classification problem. We evaluate its accuracy using Detection Error \citep{liang2017enhancing}, AUROC \citep{davis2006relationship}, and FPR at 95\% TPR \citep{liang2017enhancing}, following the methodology outlined in \citep{zhao2023open}. For action recognition tasks involving seen data, we employ Top-1 Accuracy \citep{lee2023hierarchically} for quantitative evaluation. The details of these evaluation metrics are outlined as follows:
\begin{enumerate}
    \item [1)] \textbf{Detection Error (Error)} \citep{liang2017enhancing}: It measures the misclassification probability when the True Positive Rate (TPR) is 95\%. The definition of an Error is:
    \begin{equation}
        Error=0.5\times (1-TPR)+0.5\times FPR,
    \label{error}
    \end{equation}
    where $FPR$ stands for False Positive Rate. 
    \item [2)] \textbf{AUROC} \citep{davis2006relationship}: It stands for the Area Under the Receiver Operating Characteristic (ROC) curve, which depicts the relation between TPR and FPR and ranges from 0 to 1 (the larger the value, the higher the accuracy). AUROC provides a single scalar value that summarizes the overall performance of a binary classifier across all possible thresholds.
    \item [3)] \textbf{FPR at 95\% TPR (FPR95)} \citep{liang2017enhancing}: It refers to the rate of false positive predictions when the true positive rate reaches 95\% in a classification task.
    \item [4)] \textbf{Top-1 Accuracy (ACC)} \citep{lee2023hierarchically}: It measures the proportion of correctly predicted instances among all instances, where the predicted class label exactly matches the ground truth label for each instance. It is defined as:
    \begin{equation}
        ACC=\frac{hit}{N} \times 100\%,
    \label{auroc}
    \end{equation}
    where $hit$ means the number of samples that were predicted correctly, and \textit{N} represents the total number of test samples.
    \item [5)] \textbf{Overlap}: For most energy-based OOD detection, the overlapping portion of the score distributions of ID and OOD indicates that OOD data is missed or ID data is misjudged as OOD. The smaller the overlap area of the two distributions, the greater the distinction between the two data. Therefore, we propose to use the distribution overlap area for evaluation, which can more intuitively reflect the error of OOD recognition.
\end{enumerate}

\item \textbf{Loss function}: As introduced in section \ref{subsec33}, we design a novelty energy loss to help the model distinguish OOD data during the training stage using only ID data. We set $\alpha$=0.1 and $m_{in}$=-25 in Eqn. \ref{energyloss}.

\end{itemize}

\subsubsection{Baselines}
Since most OOD detection models are post-hoc methods, we retain basic feature extraction parts and change different post-hoc methods to achieve OOD detection. We compare the proposed Skeleton-OOD with four main methods: one score measuring method MSP (\textbf{M}aximum \textbf{S}oftmax \textbf{P}robability) \citep{hendrycks17baseline} and three different feature activation methods – ReAct (\textbf{Re}ctified \textbf{Act}ivation) \citep{sun2021react}, DICE (\textbf{Di}rected \textbf{S}parisification) \citep{sun2022dice}, DICE+ReAct, ASH (\textbf{A}ctivation \textbf{Sh}aping) \citep{djurisic2022ash}. Details are shown below: 

\begin{itemize}
    \item \textbf{MSP} (\textbf{M}aximum \textbf{S}oftmax \textbf{P}robability): \citep{hendrycks17baseline} claims that out-of-distribution examples tend to have lower maximum softmax probabilities than known samples. This scoring-based method has shown wide success in the computer vision area.
    \item \textbf{ReAct} (\textbf{Re}ctified \textbf{Act}ivation) \citep{sun2021react}: A post-hoc method was proposed to reduce the overconfidence problem of the model on OOD data by truncating activations on the penultimate layer of a network to limit the effect of noise.
    \item \textbf{DICE} (\textbf{Di}rected \textbf{S}parisification) \citep{sun2022dice}: Another post-hoc method, based on the idea of denoising useless information, sorts the weights of its contribution (i.e., weight × activation). It claims that only a subset of units contributes to the in-distribution prediction results. However, it will result in non-negligible noise signals when measuring out-of-distribution data.
    \item \textbf{DICE + ReAct}: Following previous work \citep{djurisic2022ash}, we construct a baseline that filters out noise signals using both DICE and ReAct methods.
    \item \textbf{ASH} (\textbf{A}ctivation \textbf{Sh}aping) \citep{djurisic2022ash}: It consists of three different mechanisms to conduct pruning-based feature activation shaping, including ASH-P, ASH-S, and ASH-B.

\end{itemize}

\subsubsection{Ablation study}

To verify the effectiveness, we design ablation studies on the following components of Skeleton-OOD.

\begin{itemize}
    \item \textbf{GCN}: This is designed to measure the effect of the feature fusion block of Skeleton-OOD. As previously mentioned, Skeleton-OOD has two parts features: GCN-based features, which excel in classifying in-distribution (ID) samples, and ASH-based features, which aim to reduce overconfidence in predictions for OOD samples. This experiment can prove the improvement effect of ASH operation on the model OOD recognition task. 
    \item \textbf{ASH}: Based on the analysis of previous work, a pruning strategy that solely relies on thresholding may prove inadequate in certain scenarios, thus affecting the OOD recognition effect of the model. To verify whether such a problem also exists in unseen action detection, and to illustrate the role of the feature fusion module, we design such an ablation study experiment.
    \item \textbf{CE Loss}: To demonstrate the effect of the proposed energy loss function, we conduct experiments by training with conventional CE loss only.
\end{itemize}

Except for these, we also explored the results of replacing another backbone as feature extractors and another loss function for training.

\subsection{Results and discussion}
\label{subsection43}

\subsubsection{Performance comparison against baselines}
\label{baseline}
% ReAct \citep{sun2021react}
% Dice \citep{sun2022dice}
% Ash \citep{djurisic2022ash}
% MSP \citep{hendrycks17baseline}

We evaluate the proposed Skeleton-OOD method against baselines across three datasets for two tasks: (a) to distinguish OOD samples from an ID and OOD mixture dataset, and (b) to evaluate whether ID samples will be misclassified as OOD merely based on ID data. We also provide the average results for all models involved across the three datasets. 

\begin{table*}[t]
\caption{Comparing the results of the OOD recognition task with other methods on the ID-OOD mix test dataset. The best results are shown in bold.}
\vspace{5pt}
\resizebox{\textwidth}{!}{
\begin{tabular}{c|ccc|ccc|ccc|ccc}
\hline
\textbf{OOD} &
  \multicolumn{3}{c|}{\textbf{NTU60}} &
  \multicolumn{3}{c|}{\textbf{NTU120}} &
  \multicolumn{3}{c|}{\textbf{Kinetics400}} &
  \multicolumn{3}{c}{\textbf{Average}} \\ \hline
Methods &
  \multicolumn{1}{c|}{\begin{tabular}[c]{@{}c@{}}Error ↓\end{tabular}} &
  \multicolumn{1}{c|}{\begin{tabular}[c]{@{}c@{}}FPR95 ↓\end{tabular}} &
  \begin{tabular}[c]{@{}c@{}}AUROC ↑\end{tabular} &
  \multicolumn{1}{c|}{\begin{tabular}[c]{@{}c@{}}Error ↓\end{tabular}} &
  \multicolumn{1}{c|}{\begin{tabular}[c]{@{}c@{}}FPR95 ↓\end{tabular}} &
  \begin{tabular}[c]{@{}c@{}}AUROC ↑\end{tabular} &
  \multicolumn{1}{c|}{\begin{tabular}[c]{@{}c@{}}Error ↓\end{tabular}} &
  \multicolumn{1}{c|}{\begin{tabular}[c]{@{}c@{}}FPR95 ↓\end{tabular}} &
  \begin{tabular}[c]{@{}c@{}}AUROC ↑\end{tabular} &
  \multicolumn{1}{c|}{\begin{tabular}[c]{@{}c@{}}Error ↓\end{tabular}} &
  \multicolumn{1}{c|}{\begin{tabular}[c]{@{}c@{}}FPR95 ↓\end{tabular}} &
  \begin{tabular}[c]{@{}c@{}}AUROC ↑\end{tabular} \\ \hline
MSP &
  \multicolumn{1}{c|}{38.58} &
  \multicolumn{1}{c|}{72.16} &
  84.25 &
  \multicolumn{1}{c|}{50.00} &
  \multicolumn{1}{c|}{59.69} &
  86.94 &
  \multicolumn{1}{c|}{\textbf{50.00}} &
  \multicolumn{1}{c|}{92.82} &
  46.28 &
  \multicolumn{1}{c|}{46.19} &
  \multicolumn{1}{c|}{74.89} &
  72.49 \\ 
ReAct &
  \multicolumn{1}{c|}{49.99} &
  \multicolumn{1}{c|}{80.29} &
  59.79 &
  \multicolumn{1}{c|}{50.00} &
  \multicolumn{1}{c|}{99.60} &
  37.98 &
  \multicolumn{1}{c|}{51.19} &
  \multicolumn{1}{c|}{95.88} &
  47.65 &
  \multicolumn{1}{c|}{50.39} &
  \multicolumn{1}{c|}{91.92} &
  48.47 \\ 
DICE &
  \multicolumn{1}{c|}{52.47} &
  \multicolumn{1}{c|}{99.95} &
  35.62 &
  \multicolumn{1}{c|}{24.24} &
  \multicolumn{1}{c|}{56.40} &
  84.84 &
  \multicolumn{1}{c|}{51.12} &
  \multicolumn{1}{c|}{96.16} &
  46.34 &
  \multicolumn{1}{c|}{42.61} &
  \multicolumn{1}{c|}{84.17} &
  55.60 \\ 
D+R &
  \multicolumn{1}{c|}{52.49} &
  \multicolumn{1}{c|}{100.00} &
  30.80 &
  \multicolumn{1}{c|}{50.00} &
  \multicolumn{1}{c|}{99.60} &
  37.98 &
  \multicolumn{1}{c|}{51.19} &
  \multicolumn{1}{c|}{95.88} &
  47.66 &
  \multicolumn{1}{c|}{51.23} &
  \multicolumn{1}{c|}{98.49} &
  38.81 \\ 
ASH-P &
  \multicolumn{1}{c|}{\textbf{26.87}} &
  \multicolumn{1}{c|}{58.33} &
  84.48 &
  \multicolumn{1}{c|}{21.12} &
  \multicolumn{1}{c|}{56.78} &
  86.20 &
  \multicolumn{1}{c|}{50.16} &
  \multicolumn{1}{c|}{92.55} &
  46.75 &
  \multicolumn{1}{c|}{\textbf{32.72}} &
  \multicolumn{1}{c|}{69.22} &
  72.48 \\ 
ASH-S &
  \multicolumn{1}{c|}{50.00} &
  \multicolumn{1}{c|}{59.77} &
  78.73 &
  \multicolumn{1}{c|}{50.00} &
  \multicolumn{1}{c|}{60.72} &
  85.46 &
  \multicolumn{1}{c|}{51.22} &
  \multicolumn{1}{c|}{95.74} &
  48.82 &
  \multicolumn{1}{c|}{50.41} &
  \multicolumn{1}{c|}{72.08} &
  71.00 \\ 
ASH-B &
  \multicolumn{1}{c|}{42.72} &
  \multicolumn{1}{c|}{78.77} &
  68.43 &
  \multicolumn{1}{c|}{49.99} &
  \multicolumn{1}{c|}{58.53} &
  85.96 &
  \multicolumn{1}{c|}{51.11} &
  \multicolumn{1}{c|}{96.25} &
  46.72 &
  \multicolumn{1}{c|}{47.94} &
  \multicolumn{1}{c|}{77.85} &
  67.04 \\ \hline
\textbf{\begin{tabular}[c]{@{}c@{}}Ours\\ ASH-P\end{tabular}} &
  \multicolumn{1}{c|}{30.42} &
  \multicolumn{1}{c|}{\textbf{52.68}} &
  \textbf{86.64} &
  \multicolumn{1}{c|}{21.36} &
  \multicolumn{1}{c|}{60.64} &
  86.40 &
  \multicolumn{1}{c|}{51.37} &
  \multicolumn{1}{c|}{\textbf{90.61}} &
  \textbf{50.43} &
  \multicolumn{1}{c|}{34.38} &
  \multicolumn{1}{c|}{\textbf{67.98}} &
  \textbf{74.49} \\ 
\textbf{\begin{tabular}[c]{@{}c@{}}Ours\\ ASH-S\end{tabular}} &
  \multicolumn{1}{c|}{33.72} &
  \multicolumn{1}{c|}{62.45} &
  84.10 &
  \multicolumn{1}{c|}{37.87} &
  \multicolumn{1}{c|}{59.67} &
  85.09 &
  \multicolumn{1}{c|}{50.37} &
  \multicolumn{1}{c|}{93.67} &
  48.23 &
  \multicolumn{1}{c|}{40.65} &
  \multicolumn{1}{c|}{71.93} &
  72.47 \\ 
\textbf{\begin{tabular}[c]{@{}c@{}}Ours\\ ASH-B\end{tabular}} &
  \multicolumn{1}{c|}{30.21} &
  \multicolumn{1}{c|}{55.43} &
  85.98 &
  \multicolumn{1}{c|}{\textbf{19.57}} &
  \multicolumn{1}{c|}{\textbf{56.11}} &
  \textbf{87.17} &
  \multicolumn{1}{c|}{51.25} &
  \multicolumn{1}{c|}{95.62} &
  47.17 &
  \multicolumn{1}{c|}{33.68} &
  \multicolumn{1}{c|}{69.05} &
  73.44 \\ \hline
\end{tabular}}
\label{mix}
\end{table*}

As seen from Table~\ref{mix}, our method achieves the best performance in most metrics on the OOD test dataset. \textbf{On the NTU60 dataset, our method achieves an average AUROC of 85.57\% for OOD detection, with the accuracy of seen classification exceeding 91\% in Table~\ref{seen}.} Compared with several post-hoc methods, the average AUROC increases by 43.48\%. Based on the average results on the three data sets, our method achieves the best on almost all indicators (FPR95 and AUROC). To some extent, our method alleviates the problem of overconfidence in the trained model when facing OOD samples using only an end-to-end model. 

For the ID sample-only evaluation in Table~\ref{seen}, the best accuracy in ID multi-class classification may sometimes be achieved by other methods, especially in the NTU60 dataset. However, these methods show poor performance in OOD detection tasks (see Table~\ref{mix}), which means that the model still recognizes most of the data as ID. Consequently, the issue of overconfidence persists. Nevertheless, based on the average performance of the three data sets, our method still obtains the highest classification accuracy in the ID sample-only test. In summary, our method maintains the accuracy of ID data classification while also ensuring effective detection capability for OOD tasks. This phenomenon may result from ASH-P’s zero-or-not mechanism. Our analysis indicates that, with the energy-based loss function, ASH-P produces a more pronounced peak in the energy-score distribution for ID samples and a narrower range, as seen in Figure~\ref{viz-distribution} (d). In contrast, the ASH-B and ASH-S methods fill in the none-zeros with the designed ``averaged'' values, making it harder for the model to distinguish irrelevant information during training.

At the same time, by comparing the results of the three activation methods of our model horizontally, the performance of the three methods on different data sets has its advantages and disadvantages. As can be seen from Table~\ref{mix}, in the OOD detection task, ASH-P and ASH-B perform better in most indicators. When considering the average performance across all three datasets, the ASH-P activation method demonstrates better dataset transferability and greater stability. From Table~\ref{seen}, our method performs more prominently on the NTU120 and Kinetics data sets on ID data, especially under the activation of ASH-P and ASH-B methods. Judging by the number of top-performing indicators across the three datasets and the average indicator results, ASH-P remains superior. In summary, for real-world OOD detection with Skeleton-OOD, the ASH-P method is recommended for optimal results.

\begin{table*}[]
\centering
\caption{Comparing the results of the ID multi-class classification task with other methods on the ID-only test dataset. The best results are shown in bold.}
\vspace{5pt}
\resizebox{\textwidth}{!}{
\begin{tabular}{c|cc|cc|cc|cc}
\hline
\textbf{ID-only} &
  \multicolumn{2}{c|}{\textbf{NTU60}} &
  \multicolumn{2}{c|}{\textbf{NTU120}} &
  \multicolumn{2}{c|}{\textbf{Kinetics400}} &
  \multicolumn{2}{c}{\textbf{Average}} \\ \hline
Methods &
  \multicolumn{1}{c|}{Top1 ↑} &
  Overlap ↓ &
  \multicolumn{1}{c|}{Top1 ↑} &
  Overlap ↓ &
  \multicolumn{1}{c|}{Top1 ↑} &
  Overlap ↓ &
  \multicolumn{1}{c|}{Top1 ↑} &
  Overlap ↓ \\ \hline
MSP &
  \multicolumn{1}{c|}{91.69} &
  \textbf{0.35} &
  \multicolumn{1}{c|}{89.64} &
  0.41 &
  \multicolumn{1}{c|}{22.39} &
  \textbf{0.88} &
  \multicolumn{1}{c|}{67.91} &
  \textbf{0.55} \\ 
ReAct &
  \multicolumn{1}{c|}{88.27} &
  0.83 &
  \multicolumn{1}{c|}{83.82} &
  0.80 &
  \multicolumn{1}{c|}{20.98} &
  0.95 &
  \multicolumn{1}{c|}{64.36} &
  0.86 \\ 
DICE &
  \multicolumn{1}{c|}{\textbf{94.23}} &
  0.72 &
  \multicolumn{1}{c|}{76.29} &
  0.47 &
  \multicolumn{1}{c|}{21.37} &
  0.93 &
  \multicolumn{1}{c|}{63.96} &
  0.71 \\ 
D+R &
  \multicolumn{1}{c|}{92.76} &
  0.67 &
  \multicolumn{1}{c|}{47.54} &
  0.78 &
  \multicolumn{1}{c|}{20.37} &
  0.95 &
  \multicolumn{1}{c|}{53.56} &
  0.80 \\ 
ASH-P &
  \multicolumn{1}{c|}{91.32} &
  0.45 &
  \multicolumn{1}{c|}{88.91} &
  0.43 &
  \multicolumn{1}{c|}{25.18} &
  0.92 &
  \multicolumn{1}{c|}{68.47} &
  0.60 \\ 
ASH-S &
  \multicolumn{1}{c|}{91.39} &
  0.55 &
  \multicolumn{1}{c|}{89.35} &
  0.45 &
  \multicolumn{1}{c|}{23.84} &
  0.98 &
  \multicolumn{1}{c|}{68.19} &
  0.66 \\ 
ASH-B &
  \multicolumn{1}{c|}{89.69} &
  0.73 &
  \multicolumn{1}{c|}{88.98} &
  0.42 &
  \multicolumn{1}{c|}{20.81} &
  0.94 &
  \multicolumn{1}{c|}{66.49} &
  0.70 \\ \hline
\textbf{Ours ASH-P} &
  \multicolumn{1}{c|}{91.94} &
  0.39 &
  \multicolumn{1}{c|}{\textbf{89.84}} &
  0.40 &
  \multicolumn{1}{c|}{\textbf{26.18}} &
  0.89 &
  \multicolumn{1}{c|}{\textbf{69.32}} &
  0.56 \\ 
\textbf{Ours ASH-S} &
  \multicolumn{1}{c|}{91.22} &
  0.42 &
  \multicolumn{1}{c|}{88.98} &
  0.43 &
  \multicolumn{1}{c|}{23.22} &
  0.90 &
  \multicolumn{1}{c|}{67.81} &
  0.58 \\ 
\textbf{Ours ASH-B} &  
  \multicolumn{1}{c|}{91.61} &
  0.41 &
  \multicolumn{1}{c|}{89.13} &
  \textbf{0.39} &
  \multicolumn{1}{c|}{25.04} &
  0.95 &
  \multicolumn{1}{c|}{68.59} &
  0.58 \\ \hline
\end{tabular}}
\label{seen}
\end{table*}

In addition, by comparing the results of our method across the three ASH strategies with those of the pure post-hoc ASH strategy, it is evident that incorporating this feature activation operation during the training stage can enhance the model's OOD detection capability to some extent. For instance, as shown in Table~\ref{mix}, the ASH-P baseline in the fifth row compared to our Ours-ASH-P in the third to last row demonstrates varying degrees of improvement in most indicators across the three datasets. Furthermore, Table~\ref{seen} illustrates that this approach even slightly enhances the model's recognition accuracy for ID data.

It can be observed that compared to the first two datasets, the overall prediction accuracy on Kinetics is not satisfactory. Additionally, the direct classification results on the ID dataset are relatively poor. By comparing the incorrectly predicted samples on the Kinetics data set, we discover that some ID samples are misclassified as OOD samples, which means that the model lacks confidence in the ID samples. This also shows to some extent that the data itself may lack expressive power, which prevents the model itself from learning more distinguishing features. One possible reason could be attributed to the graph construction as shown in Figure~\ref{graph}. Firstly, the Kinetics dataset itself includes a large number of actions involving interactions between people and objects, making it difficult to distinguish using skeleton data only. Secondly, through our visualization analysis, there are certain issues in extracting Kinetics nodes in the STGCN \citep{yan2018spatial}. For instance, nodes 15 and 16 in the data represent the left and right eyes respectively. However, these nodes may not provide much information about the actions. Moreover, while the dataset models the left and right hips separately, it lacks crucial information about the key body waist nodes, which are essential for understanding the type of limb movement. \textbf{In summary, our method outperforms others in certain indicators for some datasets of ID samples and significantly surpasses existing methods in OOD samples. This addresses our primary challenge of enhancing OOD detection capabilities while maintaining ID recognition accuracy.}

\subsubsection{Number of dimensions for OOD representation}
Using machine learning methods to process categorical data generally requires one-hot encoding of category labels in the data. The processed vectors are then input into the model to participate in training. The dimension of the encoding vector is generally equal to the total number of data categories. In our work, to make the model realize that there are other unknown categories besides the ID category, an additional $k$-dimensional vector is concatenated to mark whether the data belongs to OOD. The experimental results on the NTU60 dataset are shown in Table \ref{dims}.

The last two rows indicate the results that we use additional dimensions ($k=1$ and $k=3$) in the embedding to represent OOD samples. Since the training set consists only of ID samples, there are no samples with these extra dimensions set to 1. We hope this operation can effectively inform the model during training that all samples belong to the first 55 classes and not to the 56th class or the 56th-58th classes, providing it with extra information. By comparing the results in the table for the rows "55", "58 (55+3)", and "56 (55+1)", it can be observed that using the additional dimension indeed improves the model's accuracy of OOD detection. This further validates the effectiveness of our design.

\begin{table}[H]
\centering
\caption{Ablation experiment results reflect the performance of each module. The bold numbers represent the best performance.}\small
\begin{tabular}{c|ccc|cc}
\hline
\multirow{2}{*}{NTU60} & \multicolumn{3}{c|}{Unseen}   & \multicolumn{2}{c}{Seen}   \\ \cline{2-6} 
 &
  \multicolumn{1}{c|}{\begin{tabular}[c]{@{}c@{}}Error ↓\end{tabular}} &
  \multicolumn{1}{c|}{\begin{tabular}[c]{@{}c@{}}FPR95 ↓\end{tabular}} &
  \begin{tabular}[c]{@{}c@{}}AUROC ↑\end{tabular} &
  \begin{tabular}[c]{@{}c@{}}Top1 ↑\end{tabular} &
  \begin{tabular}[c]{@{}c@{}}Overlap ↓\end{tabular} \\ \hline
55                     & \multicolumn{1}{c|}{34.58} & \multicolumn{1}{c|}{64.16} & 83.57 & 91.59 & 0.41 \\ \hline
58 (55+3)               & \multicolumn{1}{c|}{33.49} & \multicolumn{1}{c|}{61.99} & 85.66 & 91.49 & 0.41 \\ \hline
56 (55+1)               & \multicolumn{1}{c|}{\textbf{30.42}} & \multicolumn{1}{c|}{\textbf{52.68}} & \textbf{86.64} & \textbf{91.94} & \textbf{0.39} \\ \hline
\end{tabular}
\label{dims}
\end{table}

At the same time, comparing the results using 56(55+1) and 58(55+3) dimensions, reveals that using an extra 1 dim is more effective than that of 3 dims. Perhaps it is because higher-dimensional representations can easily cause information redundancy, which makes training difficult. Therefore, the experimental results in this paper all use additional 1-dimensional vectors to represent the OOD classes. It should be noted that perhaps it would be better to use more dimensional representations for the NTU120 and Kinetics400 datasets, but the 1-dimensional OOD representation method is used here for the requirement of comparison.

% \subsubsection{Other loss functions for OOD detection}
\subsubsection{Other loss functions for OOD detection}
\label{loss}

To evaluate the effectiveness of the proposed energy-based loss function, we compare it with another method LogitNorm \citep{wei2022mitigating}. Experiments here use our proposed workflow, with the only difference being the use of two different training loss functions. Results for both OOD detection and ID recognition tasks are shown in Table~\ref{logitnorm-mix} and Table~\ref{logitnorm-seen}, respectively.

\begin{table}[H]
\caption{Results on ID-OOD mix dataset trained with LogitNorm loss function. The best results of each column are in bold.}
\vspace{5pt}
\resizebox{\textwidth}{!}{
\begin{tabular}{c|ccc|ccc|ccc|ccc}
\hline
\textbf{OOD} &
  \multicolumn{3}{c|}{\textbf{NTU60}} &
  \multicolumn{3}{c|}{\textbf{NTU120}} &
  \multicolumn{3}{c|}{\textbf{Kinetics400}} &
  \multicolumn{3}{c}{\textbf{Average}} \\ \hline
Methods &
  \multicolumn{1}{c|}{\begin{tabular}[c]{@{}c@{}}Error ↓\end{tabular}} &
  \multicolumn{1}{c|}{\begin{tabular}[c]{@{}c@{}}FPR95 ↓\end{tabular}} &
  \begin{tabular}[c]{@{}c@{}}AUROC ↑\end{tabular} &
  \multicolumn{1}{c|}{\begin{tabular}[c]{@{}c@{}}Error ↓\end{tabular}} &
  \multicolumn{1}{c|}{\begin{tabular}[c]{@{}c@{}}FPR95 ↓\end{tabular}} &
  \begin{tabular}[c]{@{}c@{}}AUROC ↑\end{tabular} &
  \multicolumn{1}{c|}{\begin{tabular}[c]{@{}c@{}}Error ↓\end{tabular}} &
  \multicolumn{1}{c|}{\begin{tabular}[c]{@{}c@{}}FPR95 ↓\end{tabular}} &
  \begin{tabular}[c]{@{}c@{}}AUROC ↑\end{tabular} &
  \multicolumn{1}{c|}{\begin{tabular}[c]{@{}c@{}}Error ↓\end{tabular}} &
  \multicolumn{1}{c|}{\begin{tabular}[c]{@{}c@{}}FPR95 ↓\end{tabular}} &
  \begin{tabular}[c]{@{}c@{}}AUROC ↑\end{tabular} \\ \hline
\begin{tabular}[c]{@{}c@{}}LogitNorm\\ ASH-P\end{tabular} &
  \multicolumn{1}{c|}{\textbf{34.33}} &
  \multicolumn{1}{c|}{\textbf{63.66}} &
  \textbf{78.22} &
  \multicolumn{1}{c|}{\textbf{30.75}} & %%%
  \multicolumn{1}{c|}{\textbf{56.50}} &
  84.57 &
  \multicolumn{1}{c|}{\textbf{51.23}} &
  \multicolumn{1}{c|}{\textbf{88.73}} &
  52.69 &
  \multicolumn{1}{c|}{\textbf{38.77}} &
  \multicolumn{1}{c|}{\textbf{69.63}} &
  \textbf{71.83}  \\ 
\begin{tabular}[c]{@{}c@{}}LogitNorm\\ ASH-S\end{tabular} &
  \multicolumn{1}{c|}{39.11} &
  \multicolumn{1}{c|}{73.21} &
  76.50 &
  \multicolumn{1}{c|}{41.09} &
  \multicolumn{1}{c|}{77.19} &
  75.01 &
  \multicolumn{1}{c|}{51.39} &
  \multicolumn{1}{c|}{96.74} &
  46.55 &
  \multicolumn{1}{c|}{43.86} &
  \multicolumn{1}{c|}{82.38} &
  66.02 \\ 
\begin{tabular}[c]{@{}c@{}}LogitNorm\\ ASH-B\end{tabular} &
  \multicolumn{1}{c|}{43.10} &
  \multicolumn{1}{c|}{81.21} &
  75.05 &
  \multicolumn{1}{c|}{32.85} &
  \multicolumn{1}{c|}{60.71} &
  \textbf{86.70} &
  \multicolumn{1}{c|}{51.46} &
  \multicolumn{1}{c|}{94.69} &
  \textbf{52.70} &
  \multicolumn{1}{c|}{51.09} &
  \multicolumn{1}{c|}{96.10} &
  51.31 \\ 
 \hline
\end{tabular}}
\label{logitnorm-mix}
\end{table}

\begin{table}[H]
\centering
\caption{Results on ID-only test dataset trained with LogitNorm loss function. The best results of each column are in bold.}
\vspace{5pt}
\resizebox{\textwidth}{!}{
\begin{tabular}{c|cc|cc|cc|cc}
\hline
\textbf{ID-only} &
  \multicolumn{2}{c|}{\textbf{NTU60}} &
  \multicolumn{2}{c|}{\textbf{NTU120}} &
  \multicolumn{2}{c|}{\textbf{Kinetics400}} &
  \multicolumn{2}{c}{\textbf{Average}} \\ \hline
Methods &
  \multicolumn{1}{c|}{Top1 ↑} &
  Overlap ↓ &
  \multicolumn{1}{c|}{Top1 ↑} &
  Overlap ↓ &
  \multicolumn{1}{c|}{Top1 ↑} &
  Overlap ↓ &
  \multicolumn{1}{c|}{Top1 ↑} &
  Overlap ↓ \\ \hline
LogitNorm ASH-P &
  \multicolumn{1}{c|}{\textbf{90.66}} &
  \textbf{0.37} &
  \multicolumn{1}{c|}{\textbf{89.05}} &
  \textbf{0.47} &
  \multicolumn{1}{c|}{\textbf{26.30}} &
  0.92 &
  \multicolumn{1}{c|}{\textbf{68.67}} &
  \textbf{0.59} \\ \hline
LogitNorm ASH-S &
  \multicolumn{1}{c|}{90.39} &
  0.59 &
  \multicolumn{1}{c|}{88.17} &
  0.64 &
  \multicolumn{1}{c|}{21.39} &
  0.93 &
  \multicolumn{1}{c|}{66.76} &
  0.65 \\ \hline
LogitNorm ASH-B &
  \multicolumn{1}{c|}{4.44} &
  0.79  &
  \multicolumn{1}{c|}{17.63} &
  0.79 &
  \multicolumn{1}{c|}{26.19} &
  \textbf{0.91} &
  \multicolumn{1}{c|}{16.09} &
  0.83  \\ \hline
\end{tabular}}
\label{logitnorm-seen}
\end{table}

By comparison, it can be seen that in most experiments, the model trained with our proposed energy-based loss function outperforms the results obtained with LogitNorm. It can also be observed that the results trained with LogitNorm show significant variation among different activation methods, particularly with the ASH-B method. We speculate that this may be because the LogitNorm method itself is a correction approach based on the original softmax prediction activation. Moreover, our designed framework partly uses ASH to activate the skeleton features and then applies the LogitNorm loss function for correction. This may lead to either insignificant effects (as seen in LogitNorm-ASH-P and LogitNorm-ASH-S) or overcorrection (as seen in LogitNorm-ASH-B). On the other hand, our framework utilizes ASH for feature activation, and during the training stage, the proposed loss function directly reinforces the learning of the OOD discrimination target through energy distance. The experimental results also show that for our framework, using this target-enhancing training method during the training stage through loss function is more effective. Therefore, overall, our framework not only addresses the issue of OOD feature extraction from the perspective of feature activation but also strengthens the range of ID sample energy scores from the perspective of the energy value distribution with the aim of discrimination. This multi-angle approach enables the model to address both OOD and ID sample recognition problems effectively.

\subsubsection{Ablation study}

\begin{table}[H]
\centering
\caption{Ablation experiment results reflect the performance of each module. The bold numbers represent the best performance.}\small
\vspace{5pt}
\begin{tabular}{c|ccc|cc}
\hline
\multirow{2}{*}{NTU60}                                  & \multicolumn{3}{c|}{OOD}                                     & \multicolumn{2}{c}{ID}  \\ \cline{2-6} 
 &
  \multicolumn{1}{c|}{\begin{tabular}[c]{@{}c@{}}Error ↓\end{tabular}} &
  \multicolumn{1}{c|}{\begin{tabular}[c]{@{}c@{}}FPR95 ↓\end{tabular}} &
  \begin{tabular}[c]{@{}c@{}}AUROC ↑\end{tabular} &
  \multicolumn{1}{c|}{\begin{tabular}[c]{@{}c@{}}Top1 ↑\end{tabular}} &
  \begin{tabular}[c]{@{}c@{}}Overlap ↓\end{tabular} \\ \hline
GCN                                                & \multicolumn{1}{c|}{49.31} & \multicolumn{1}{c|}{93.61} & 48.37 & \multicolumn{1}{c|}{7.24}  & 0.89 \\ \hline
ASH                                                & \multicolumn{1}{c|}{34.33} & \multicolumn{1}{c|}{63.66} & 78.22 & \multicolumn{1}{c|}{90.66} & 0.59 \\ \hline
\begin{tabular}[c]{@{}c@{}}CE Loss\end{tabular} & \multicolumn{1}{c|}{37.51} & \multicolumn{1}{c|}{70.02} & 79.58 & \multicolumn{1}{c|}{91.61} & 0.49 \\ \hline
ours                                                    & \multicolumn{1}{c|}{\textbf{30.42}} & \multicolumn{1}{c|}{\textbf{52.68}} & \textbf{86.64} & \multicolumn{1}{c|}{\textbf{91.94}} & \textbf{0.39} \\ \hline
\end{tabular}
\end{table}

\begin{itemize}
    \item \textbf{Effect of Activation Shaping}: Comparing the results of the GCN with ours, it shows that with the help of the ASH strategy, the Error metric decreases by 58.59\%. AUROC accuracy also improves by 79.12\%. ASH can significantly improve the OOD recognition ability of the model. At the same time, it can be seen from the table that the seen accuracy in the GCN experiment is very low because the energy score distribution of unseen and seen has a high degree of overlap. Therefore, many ID samples will be misidentified as OOD. The OOD detection method that uses the energy score of the ranked 10\% ID sample as the threshold in OOD detection is suspected of overcorrecting the performance of this model.

    \item \textbf{Effect of Feature Fusion Block}: Comparing it to the column ASH, leveraging typical GCN embeddings results in an 11.39\% increase in the Error metric and an 8.42 percentage increase in AUROC accuracy. When considering the experimental outcomes from the GCN scenario, it becomes evident that the feature fusion module empowers the model to increase the accuracy of out-of-distribution (OOD) detection without compromising its ability to recognize in-distribution (ID) samples.

\end{itemize}

\begin{figure}[H]
    \centering
    \setlength{\abovecaptionskip}{0.cm}
    \includegraphics[width=\linewidth,scale=0.9]{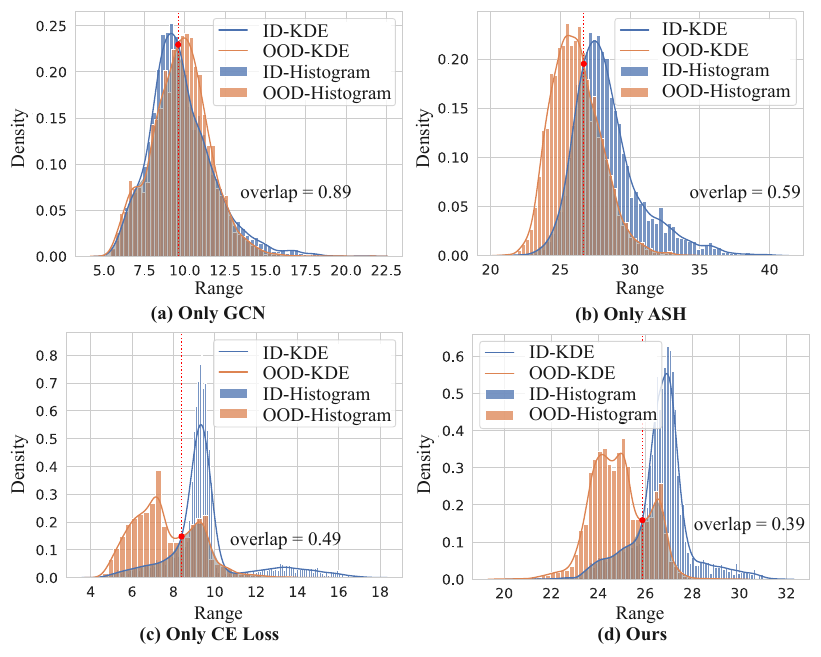}
    \caption{Distributions of ID and OOD samples among different ablation studies. The blue area indicates the ID distribution, and the orange shows the OOD samples. The value of overlap represents the amount of overlapping area between the two distributions.}
    \label{viz-distribution}
\end{figure}

\begin{itemize}
    \item \textbf{Effect of Energy Loss}: Comparing the results with only traditional cross-entropy loss (bottom two rows), it can be observed that the method utilizing energy loss employed in this paper significantly enhances the model's capabilities across various settings. Additionally, we discover that compared with directly employing ASH, the model trained with the feature fusion module but using only cross-entropy loss performs less effectively on OOD tasks, but slightly outperforms in ID classification. This further indicates that the combined use of the feature fusion module and energy loss can address the overall capabilities of the model, leading to better differentiation between in-distribution and out-of-distribution data.
    
    \item \textbf{The impact of different model components on energy score distribution}: From Figure~\ref{viz-distribution}, it can be observed that incorporating the ASH method during the training stage (Figure~\ref{viz-distribution} (b)) reduces the range of energy score distribution compared to the pure GCN model. Additionally, it creates the distance between the two distribution peaks. Contrasting Figure~\ref{viz-distribution} (b) with Figure~\ref{viz-distribution} (d), where the only difference lies in the usage of a feature fusion block, shows a further reduction in the distribution range from 20 to 12. Moreover, the variance of the ID data is also diminished. Furthermore, as evident from Figure~\ref{viz-distribution} (c) and (d), adjustments to the Energy-based loss function result in not only a decreased range of the data but also a reduction in the variance of the OOD data. The combination of these effects minimizes the overlap area of the ID and OOD distributions in our model test results.
\end{itemize}

\subsubsection{Different pruning percentages}
\label{pruning}

\begin{figure}[H]
  \centering
  \setlength{\abovecaptionskip}{0.cm}
  \includegraphics[width=\linewidth,scale=0.7]{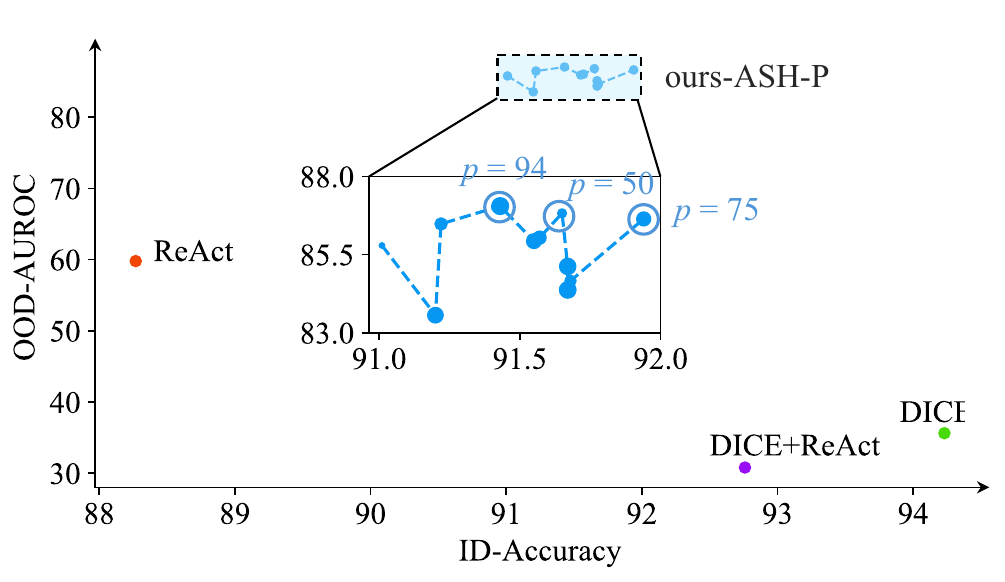}
  \caption{ID-OOD tradeoff on the NTU-RGB+D 60 dataset. The size of the points in the enlarged line chart corresponds to different $p$ values. The line chart represents the results of our method for p-values in the range of 40\%-90\%.}
  \label{prun}
\end{figure}

To explore the effect of pruning percentage $p$, different values are tested in our experiments. As shown in Figure~\ref{prun}, we select the best-performing ASH-P strategy on the NTU60 dataset for evaluation. The size of the blue points reflects the pruning percentage $p$. The red, green, and purple points respectively represent the performance of the ReAct, DICE, and DICE+ReAct methods at their optimal ratios. 

The data points located closer to the upper-right corner of the figure exhibit superior performance. Our approach achieves optimal results when integrated with the ASH strategy. Concerning the OOD detection task across various $p$ values, it is apparent that the correlation between accuracy and $p$ value is nonlinear. To balance the accuracy between OOD and ID samples, we choose the p-value of 75 as the best pruning percentage.

%% DICE effectively decreases the output variance for OOD data, resulting in a sharper score distribution and enhanced separability from ID data. Thus, we wonder why Skeleton-OOD achieves such an effective separation of ID and OOD samples using energy scores. To this end, we plotted the output distributions of various ablation study models.

\subsubsection{Different backbones of skeleton feature extractor}
\label{backbone}

As stated in Section~\ref{introduction}, we focus on exploiting the extracted features to extend their adaptability for OOD detection while maintaining robust prediction performance for the supervised classes. Thus, we use the state-of-the-art supervised skeleton-based recognition model HD-GCN as a feature extractor. Additionally, we also select ST-GCN, InfoGCN \citep{Chi_2022_CVPR} and EfficientGCN \citep{song2022constructing}, which are similar frameworks as HD-GCN in classic action recognition tasks, to serve as a feature extractor for comparative experiments. The results demonstrate that the model with HD-GCN still outperforms many other baselines.

\begin{table*}[]
\setlength{\belowcaptionskip}{8pt}
\caption{Results on ID-OOD mix dataset with three different extractors. The best results of each method are in bold. ``+'' indicates an increase in the corresponding indicator value, while ``-'' indicates a decrease.}
\resizebox{\textwidth}{!}{
\begin{tabular}{c|ccc|ccc|ccc}
\hline
\textbf{OOD} &
  \multicolumn{3}{c|}{\textbf{NTU60}} &
  \multicolumn{3}{c|}{\textbf{NTU120}} &
  \multicolumn{3}{c}{\textbf{Kinetics400}} \\ \hline
Methods &
  \multicolumn{1}{c|}{\begin{tabular}[c]{@{}c@{}}Error ↓\end{tabular}} &
  \multicolumn{1}{c|}{\begin{tabular}[c]{@{}c@{}}FPR95 ↓\end{tabular}} &
  \begin{tabular}[c]{@{}c@{}}AUROC ↑\end{tabular} &
  \multicolumn{1}{c|}{\begin{tabular}[c]{@{}c@{}}Error ↓\end{tabular}} &
  \multicolumn{1}{c|}{\begin{tabular}[c]{@{}c@{}}FPR95 ↓\end{tabular}} &
  \begin{tabular}[c]{@{}c@{}}AUROC ↑\end{tabular} &
  \multicolumn{1}{c|}{\begin{tabular}[c]{@{}c@{}}Error ↓\end{tabular}} &
  \multicolumn{1}{c|}{\begin{tabular}[c]{@{}c@{}}FPR95 ↓\end{tabular}} &
  \begin{tabular}[c]{@{}c@{}}AUROC ↑\end{tabular}  \\ \hline
  \begin{tabular}[c]{@{}c@{}}ST-GCN-only\end{tabular} &
  \multicolumn{1}{c|}{35.00} &
  \multicolumn{1}{c|}{65.01} &
  81.16 &
  \multicolumn{1}{c|}{56.07} &
  \multicolumn{1}{c|}{68.36} &
  53.85 &
  \multicolumn{1}{c|}{\textbf{51.37}} &
  \multicolumn{1}{c|}{95.11} &
  51.79 \\ 
  \begin{tabular}[c]{@{}c@{}}InfoGCN-only\end{tabular} &
  \multicolumn{1}{c|}{34.43} &
  \multicolumn{1}{c|}{66.80} &
  80.29 &
  \multicolumn{1}{c|}{52.39} &
  \multicolumn{1}{c|}{66.94} &
  59.07 &
  \multicolumn{1}{c|}{-} &
  \multicolumn{1}{c|}{-} &
  - \\ 
  \begin{tabular}[c]{@{}c@{}}EfficientGCN-only\end{tabular} &
  \multicolumn{1}{c|}{35.84} &
  \multicolumn{1}{c|}{\textbf{67.02}} &
  79.86 &
  \multicolumn{1}{c|}{54.92} &
  \multicolumn{1}{c|}{67.82} &
  58.92 &
  \multicolumn{1}{c|}{-} &
  \multicolumn{1}{c|}{-} &
  - \\\hline
\begin{tabular}[c]{@{}c@{}}ST-GCN ASH-P\end{tabular} &
  \multicolumn{1}{c|}{\textbf{-2.53}} &
  \multicolumn{1}{c|}{\textbf{-5.06}} &
  \textbf{+18.08} &
  \multicolumn{1}{c|}{-4.39} &
  \multicolumn{1}{c|}{+3.13} &
  +18.08 &
  \multicolumn{1}{c|}{+0.53} &
  \multicolumn{1}{c|}{-2.38} &
  +5.10 \\ 
\begin{tabular}[c]{@{}c@{}}ST-GCN ASH-S\end{tabular} &
  \multicolumn{1}{c|}{+1.83} &
  \multicolumn{1}{c|}{+3.66} &
  -4.53 &
  \multicolumn{1}{c|}{-23.11} &
  \multicolumn{1}{c|}{-7.43} &
  +29.05 &
  \multicolumn{1}{c|}{+0.45} &
  \multicolumn{1}{c|}{-2.02} &
  +3.06  \\ 
\begin{tabular}[c]{@{}c@{}}ST-GCN ASH-B\end{tabular} &
  \multicolumn{1}{c|}{+8.10} &
  \multicolumn{1}{c|}{+16.20} &
  -6.11 &
  \multicolumn{1}{c|}{\textbf{-23.22}} &
  \multicolumn{1}{c|}{\textbf{-7.65}} &
  \textbf{+32.85} &
  \multicolumn{1}{c|}{+0.55} &
  \multicolumn{1}{c|}{\textbf{-2.44}} &
  \textbf{+5.29}  \\ \hline
  \begin{tabular}[c]{@{}c@{}}InfoGCN ASH-P\end{tabular} &
  \multicolumn{1}{c|}{\textbf{-1.08}} &
  \multicolumn{1}{c|}{\textbf{-0.88}} &
  \textbf{+4.60} &
  \multicolumn{1}{c|}{\textbf{-23.22}} &
  \multicolumn{1}{c|}{+1.44} &
  \textbf{+28.65} &
  \multicolumn{1}{c|}{-} &
  \multicolumn{1}{c|}{-} &
  - \\ 
\begin{tabular}[c]{@{}c@{}}InfoGCN ASH-S\end{tabular} &
  \multicolumn{1}{c|}{-0.23} &
  \multicolumn{1}{c|}{-0.06} &
  +2.04 &
  \multicolumn{1}{c|}{-9.02} &
  \multicolumn{1}{c|}{+3.34} &
  +20.75 &
  \multicolumn{1}{c|}{-} &
  \multicolumn{1}{c|}{-} &
  -  \\ 
\begin{tabular}[c]{@{}c@{}}InfoGCN ASH-B\end{tabular} &
  \multicolumn{1}{c|}{+0.36} &
  \multicolumn{1}{c|}{+1.18} &
  -3.36 &
  \multicolumn{1}{c|}{-11.52} &
  \multicolumn{1}{c|}{\textbf{-0.02}} &
  +24.49 &
  \multicolumn{1}{c|}{-} &
  \multicolumn{1}{c|}{-} &
  -  \\ \hline
  \begin{tabular}[c]{@{}c@{}}EfficientGCN ASH-P\end{tabular} &
  \multicolumn{1}{c|}{+4.44} &
  \multicolumn{1}{c|}{+5.35} &
  \textbf{+1.46} &
  \multicolumn{1}{c|}{-11.00} &
  \multicolumn{1}{c|}{+0.56} &
  +22.06 &
  \multicolumn{1}{c|}{-} &
  \multicolumn{1}{c|}{-} &
  - \\ 
\begin{tabular}[c]{@{}c@{}}EfficientGCN ASH-S\end{tabular} &
  \multicolumn{1}{c|}{\textbf{-1.02}} &
  \multicolumn{1}{c|}{+1.90} &
  +0.87 &
  \multicolumn{1}{c|}{-15.63} &
  \multicolumn{1}{c|}{-3.53} &
  +22.28 &
  \multicolumn{1}{c|}{-} &
  \multicolumn{1}{c|}{-} &
  -  \\ 
\begin{tabular}[c]{@{}c@{}}EfficientGCN ASH-B\end{tabular} &
  \multicolumn{1}{c|}{+7.05} &
  \multicolumn{1}{c|}{+11.50} &
  -1.12 &
  \multicolumn{1}{c|}{\textbf{-16.63}} &
  \multicolumn{1}{c|}{\textbf{-3.99}} &
  \textbf{+23.27} &
  \multicolumn{1}{c|}{-} &
  \multicolumn{1}{c|}{-} &
  -  \\ \hline
\end{tabular}}
\label{stgcn-mix}
\end{table*}

\begin{table}[H]
\centering
\setlength{\belowcaptionskip}{8pt}
\caption{Results on ID-only dataset with three different extractors. The best results of each method are in bold. ``+'' indicates an increase in the corresponding indicator value, while ``-'' indicates a decrease.}
\resizebox{\textwidth}{!}{
\footnotesize
\begin{tabular}{c|cc|cc|cc}
\hline
\textbf{ID-only} &
  \multicolumn{2}{c|}{\textbf{NTU60}} &
  \multicolumn{2}{c|}{\textbf{NTU120}} &
  \multicolumn{2}{c}{\textbf{Kinetics400}} \\ \hline
Methods &
  \multicolumn{1}{c|}{Top1 ↑} &
  Overlap ↓ &
  \multicolumn{1}{c|}{Top1 ↑} &
  Overlap ↓ &
  \multicolumn{1}{c|}{Top1 ↑} &
  Overlap ↓ \\ \hline
  ST-GCN-only &
  \multicolumn{1}{c|}{90.33} &
  0.53 &
  \multicolumn{1}{c|}{48.11} &
  0.81 &
  \multicolumn{1}{c|}{\textbf{24.03}} &
  0.92  \\
  InfoGCN-only &
  \multicolumn{1}{c|}{90.95} &
  0.51  &
  \multicolumn{1}{c|}{73.64} &
  0.76  &
  \multicolumn{1}{c|}{-} &
  -  \\
  EfficientGCN-only &
  \multicolumn{1}{c|}{90.88} &
  0.55 &
  \multicolumn{1}{c|}{63.21} &
  0.79 &
  \multicolumn{1}{c|}{-} &
  -  \\\hline
ST-GCN ASH-P &
  \multicolumn{1}{c|}{\textbf{+0.64}} &
  \textbf{-0.10} &
  \multicolumn{1}{c|}{\textbf{+40.88}} &
  -0.09 &
  \multicolumn{1}{c|}{-2.10} &
  \textbf{-0.03} \\ 
ST-GCN ASH-S &
  \multicolumn{1}{c|}{-0.25} &
  +0.03 &
  \multicolumn{1}{c|}{+40.70} &
  -0.35 &
  \multicolumn{1}{c|}{-2.64} &
  +0.01 \\ 
ST-GCN ASH-B &
  \multicolumn{1}{c|}{+0.08} &
  +0.08 &
  \multicolumn{1}{c|}{+39.86} &
  \textbf{-0.40} &
  \multicolumn{1}{c|}{-2.49} &
  \textbf{-0.03} \\ \hline
  InfoGCN ASH-P &
  \multicolumn{1}{c|}{\textbf{+0.92}} &
  \textbf{-0.04} &
  \multicolumn{1}{c|}{+15.30} &
  -0.08 &
  \multicolumn{1}{c|}{-} &
  - \\ 
InfoGCN ASH-S &
  \multicolumn{1}{c|}{-0.01} &
  -0.01 &
  \multicolumn{1}{c|}{+15.38} &
  -0.08 &
  \multicolumn{1}{c|}{-} &
  - \\ 
InfoGCN ASH-B &
  \multicolumn{1}{c|}{-0.20} &
  +0.07 &
  \multicolumn{1}{c|}{\textbf{+15.60}} &
  \textbf{-0.23} &
  \multicolumn{1}{c|}{-} &
  - \\ \hline

  EfficientGCN ASH-P &
  \multicolumn{1}{c|}{\textbf{+0.35}} &
  \textbf{-0.13} &
  \multicolumn{1}{c|}{+24.71} &
  \textbf{-0.32} &
  \multicolumn{1}{c|}{-} &
  - \\ 
EfficientGCN ASH-S &
  \multicolumn{1}{c|}{+0.14} &
  -0.11 &
  \multicolumn{1}{c|}{+25.22} &
  -0.27 &
  \multicolumn{1}{c|}{-} &
  - \\ 
EfficientGCN ASH-B &
  \multicolumn{1}{c|}{-0.09} &
  -0.02 &
  \multicolumn{1}{c|}{\textbf{+25.51}} &
  -0.22 &
  \multicolumn{1}{c|}{-} &
  - \\ \hline
\end{tabular}}
\label{stgcn-seen}
\end{table}

It should be noted that we use the Kinetics400 dataset processed by the work ST-GCN \citep{yan2018spatial}, while the model InfoGCN and EfficientGCN are designed for motion stream data, so the results on this dataset are missing. 

Compared with the results in Table~\ref{mix} and Table~\ref{stgcn-mix}, the three feature extractor backbones perform less effectively than Skeleton-OOD on the either OOD or ID recognition task. This result indicates that HD-GCN has a stronger capability as a feature extractor than these three. This aligns with findings from numerous existing studies in the field of action recognition, which show that HD-GCN holds the state-of-the-art performance \citep{lee2023hierarchically}. This further confirms that a superior feature extractor significantly aids in improving OOD recognition. The reason for this phenomenon may not only be related to the design of the network structure, but more importantly, the graph structures used by the methods are different. 

Also, comparing the results of each backbone and the below activation-added methods in Table~\ref{stgcn-mix} and Table~\ref{stgcn-seen}, it can be concluded that adding activation operations after the feature extractor can indeed improve the OOD detection capability of the model. This is not only limited to HD-GCN as an extractor but also works on models with other structures.

\subsubsection{Detection result analysis and visualization}
Based on the performance of the train and test sets of NTU-RGB+D 60, the energy distribution in the test set is bimodal. We suspect that some out-of-distribution classes are similar to in-distribution classes, making them difficult to distinguish. We list these `confusing' classes in Table~\ref{categories}.

\begin{table}[!ht]
    \caption{Misclassification statistics}
    \resizebox{\textwidth}{!}{
    \begin{tabular}{c|c|c|c|c|c|c}
    \hline
        Wrong number & 533 & 73 & 48 & 93 & 96 & 50 \\ \hline
        ID classes & clapping & giving object & wipe face & headache & fan self & pat on back \\ \hline
        OOD classes & \multicolumn{6}{l}{hand waving, kicking something, rubbing two hands, neck pain, touch pocket} \\ \hline
    \end{tabular}}
    \label{categories}
\end{table}

\begin{figure}[H]
\centering
\subfigure[A10-ID: clapping]{\includegraphics[width=0.45\textwidth]{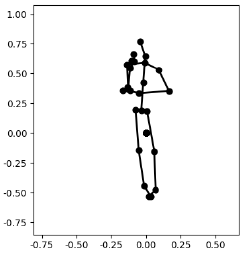}} \hfill
\subfigure[A23-OOD: hand waving]{\includegraphics[width=0.47\textwidth]{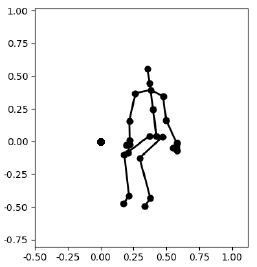}} \\
\subfigure[A53-ID: pat on back]{\includegraphics[width=0.45\textwidth]{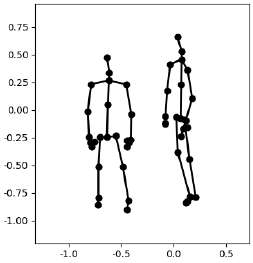}} \hfill
\subfigure[A47-OOD: neck pain]{\includegraphics[width=0.46\textwidth]{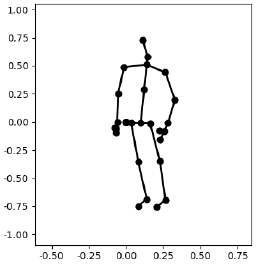}}
\caption{Example of correct classified ID samples and misclassified OOD samples.}
\label{viz-fig}
\end{figure}

Table~\ref{categories} lists out the count of samples and their class names respectively from OOD samples that are misclassified into ID classes. Upon examining the class names, it is apparent that certain actions have subtle distinctions but are still similar, such as `clapping' and `hand waving', or `pat on back' and `neck pain' as shown in Figure~\ref{viz-fig}. How the model captures the little differences between these actions still needs some improvement.

\subsubsection{Randomly joint masking result}

\begin{figure}[H]
  \centering
  \setlength{\abovecaptionskip}{0.cm}
  \includegraphics[width=\linewidth,scale=0.6]{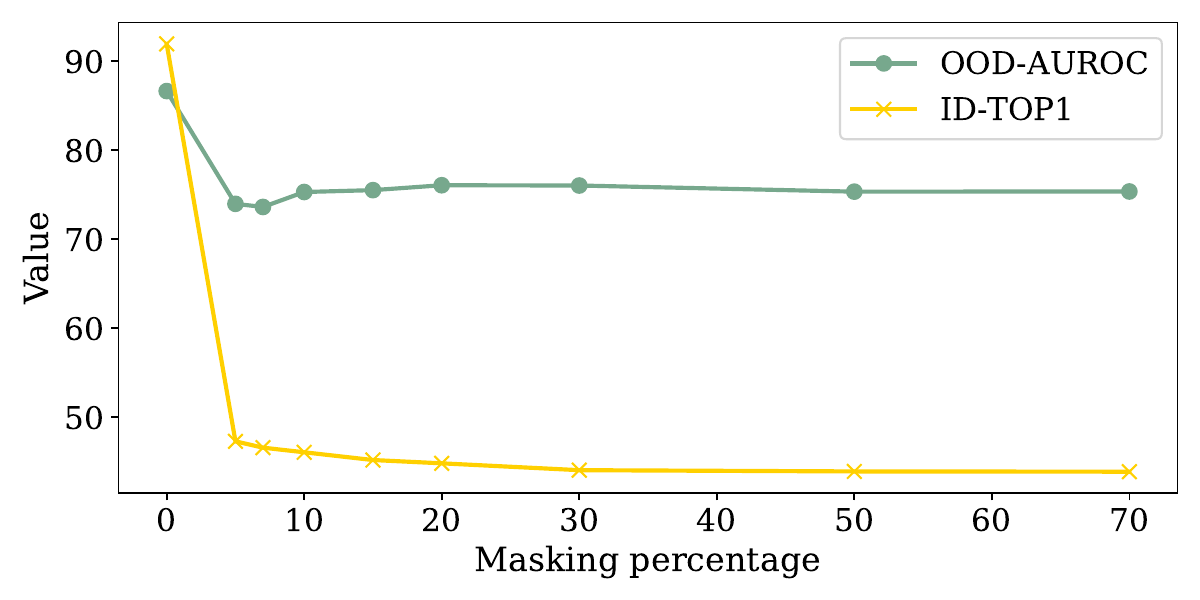}
  \caption{Results of ID and OOD performance under different masking percentages.}
  \label{line-mask}
\end{figure}

In order to test the scalability of the model on multiple tasks, we also considered the results of the recognition and classification problems with missing node information in reality. The results of ID and OOD detection on the NTU60 data are shown in Table~\ref{mask}. And plot the trend of AUROC and TOP1 results as shown in Figure~\ref{line-mask}.

\begin{table}[!ht]
    \centering
    \setlength{\belowcaptionskip}{10pt} % 设置标题下与表格之间的间距
    \caption{Test results of different joint mask percentages (\textit{p}). Using our model (ASH-P), we randomly masked different percentages of joint information on the NTU-RGB+D 60 ID-OOD mix dataset.}
    \footnotesize % 设置字体为小号
    \begin{tabular}{c|ccc|cc}
    \hline
        \multirow{2}{*}{} & \multicolumn{3}{c|}{OOD} & \multicolumn{2}{c}{ID} \\ \cline{2-6}
        &  \multicolumn{1}{c|}{ERROR ↓} & \multicolumn{1}{c|}{FPR95 ↓} & AUROC ↑ & \multicolumn{1}{c|}{Top1 ↑} & Overlap ↓ \\ \hline
        origin(0) & \multicolumn{1}{c|}{30.42} & \multicolumn{1}{c|}{52.68} & \multicolumn{1}{c|}{86.64} & \multicolumn{1}{c|}{91.94} & \multicolumn{1}{c}{0.39}  \\ 
        \textit{p}=5\% & \multicolumn{1}{c|}{33.99} & \multicolumn{1}{c|}{52.32} & \multicolumn{1}{c|}{73.97} & \multicolumn{1}{c|}{47.29} & \multicolumn{1}{c}{0.46}  \\ 
        \textit{p}=7\% & \multicolumn{1}{c|}{33.56} & \multicolumn{1}{c|}{53.27} & \multicolumn{1}{c|}{73.62} & \multicolumn{1}{c|}{46.58} & \multicolumn{1}{c}{0.46}  \\ 
        \textit{p}=10\% & \multicolumn{1}{c|}{33.74} & \multicolumn{1}{c|}{52.86} & \multicolumn{1}{c|}{75.30} & \multicolumn{1}{c|}{46.06} & \multicolumn{1}{c}{0.49}  \\ 
        \textit{p}=15\% & \multicolumn{1}{c|}{33.72} & \multicolumn{1}{c|}{52.91} & \multicolumn{1}{c|}{75.51} & \multicolumn{1}{c|}{45.19} & \multicolumn{1}{c}{0.49}  \\ 
        \textit{p}=20\% & \multicolumn{1}{c|}{33.98} & \multicolumn{1}{c|}{52.33} & \multicolumn{1}{c|}{76.06} & \multicolumn{1}{c|}{44.82} & \multicolumn{1}{c}{0.49}  \\ 
        \textit{p}=30\% & \multicolumn{1}{c|}{34.18} & \multicolumn{1}{c|}{51.90} & \multicolumn{1}{c|}{76.03} & \multicolumn{1}{c|}{44.05} & \multicolumn{1}{c}{0.49}  \\ 
        \textit{p}=50\% & \multicolumn{1}{c|}{34.11} & \multicolumn{1}{c|}{52.04} & \multicolumn{1}{c|}{75.34} & \multicolumn{1}{c|}{43.91} & \multicolumn{1}{c}{0.49}  \\ 
        \textit{p}=70\% & \multicolumn{1}{c|}{34.32} & \multicolumn{1}{c|}{51.60} & \multicolumn{1}{c|}{75.36} & \multicolumn{1}{c|}{43.88} & \multicolumn{1}{c}{0.49} \\ \hline
    \end{tabular}
    \label{mask}
\end{table}

From the results we can clearly see that randomly masking on joints affects the accuracy of the ID data most, but slightly on OOD samples. Especially when the percentage is small, the OOD result is not just a simple downward trend. This may be related to the position of the nodes randomly masked.

\section{Conclusion and future works}
\label{conclusion}

This paper introduces Skeleton-OOD, an end-to-end framework designed to address the challenge of recognizing out-of-distribution human action samples. The framework adopts a hierarchical graph composition method and utilizes GCN to extract the feature maps of human skeletons. Subsequently, a feature activation strategy is applied and we fuse the activated features with the original ones through a proposed feature fusion block. Experiments show that this step equips the model with the innate ability to identify out-of-distribution data while maintaining its capability for ID classification. Next, an energy score is computed for each test sample based on the logits output by the classifier. This helps determine whether it belongs to an OOD category. The model is trained using a designed energy-based loss function, designed to reduce the variance of ID data and differentiate between ID and OOD samples. Experimental results demonstrate that this approach effectively mitigates the overconfidence issue encountered by human action classification models when dealing with samples beyond the training category. Notably, since the model training process relies exclusively on sample information within the branch, without prior knowledge of OOD distribution characteristics, it exhibits broad practical applicability.

Admittedly, the model still has some potential limitations. Firstly, since the model is trained on GCN, its generalization ability is constrained by the number of skeleton nodes across various datasets. In real-world scenarios, if the trained model is deployed to identify whether a test sample with a different number of nodes is out-of-distribution, its accuracy may be compromised. Secondly, the end-to-end model's generalization capability is still bounded by the model parameters and the size of the dataset. The way the activation method is used during training causes some of the improvement in results to come from the training operation. One potential solution is to explore semi-supervised learning techniques to mitigate this issue. Moreover, the problem can be further addressed by introducing generative models and conducting training on datasets with limited ID samples. By leveraging existing category text and feature information, the model can infer the name of the unknown category for unfamiliar sample features. In summary, human action recognition tasks present diverse application scenarios, with numerous novel challenges awaiting exploration based on specific practical contexts.

\appendix
\section{Appendix}
\label{app1}
\subsection{Three different activation shaping strategies}
In our paper, we try three different activation shaping strategies and compare the results respectively. Here are their implementation details in Table~\ref{activate-alg}.

\begin{table}[!ht]
    % \centering
    \caption{Detailed activation shaping algorithms}
    \resizebox{\textwidth}{!}{
    \begin{tabular}{l|l|l}
    \hline
        \multicolumn{3}{l}{Algorithm S1. Activation shaping algorithms} \\ \hline
        \multicolumn{3}{l}{\textbf{Input}:\quad Feature maps $F$;} \\ 
        \multicolumn{3}{l}{\qquad\qquad Pruning percentage $p$;} \\ 
        \multicolumn{3}{l}{\textbf{Output}: Modified feature maps $F_P$, $F_B$, or $F_S$;} \\ \hline
        ASH-P: Activation Shaping with Pruning & ASH-B: Activation Shaping with Binarizing & ASH-S: Activation Shaping with Scaling \\ \hline
        1: $t=p \times \operatorname{rank}(F)$ & 1: $t=p \times \operatorname{rank}(F)$ & 1: $t=p \times \operatorname{rank}(F)$ \\ 
        2: $F_P[\operatorname{idx}(\mathrm{F}<\mathrm{t})]=0$ & 2: $s=\operatorname{sum}(F)$ & 2: $s1=\operatorname{sum}(F)$ \\ 
        3: return $F_P$ & 3: $F_P[\operatorname{idx}(\mathrm{F}<\mathrm{t})]=0$ & 3: $F_P[\operatorname{idx}(\mathrm{F}<\mathrm{t})]=0$ \\ 
        ~ & 4: $n=\operatorname{len}(\operatorname{idx}(F \neq 0))$ & 4: $s2=\operatorname{sum}(F)$ \\ 
        ~ & 5: $F_B[F \neq 0]=s / n$ & 5: $F_S=F[\operatorname{idx}(F \neq 0)] \times \exp (s 1 / s 2)$ \\ 
        ~ & 6: return $F_B$ & 6: return $F_S$ \\ \hline
    \end{tabular}}
    \label{activate-alg}
\end{table}

%%       \citet{<label>} ==> Lamport [21]
%%       \citep{<label>} ==> [21]
%%
%% Example citation, See \citet{lamport94}.

\bibliographystyle{elsarticle-num-names} 
\bibliography{refs}



%% \begin{thebibliography}{00}


%% \bibitem[Lamport(1994)]{lamport94}
%%   Leslie Lamport,
%%   \textit{\LaTeX: a document preparation system},
%%   Addison Wesley, Massachusetts,
%%   2nd edition,
%%   1994.

%% \end{thebibliography}
\end{document}